\def\year{2017}\relax
\documentclass[letterpaper]{article}
\usepackage{aaai17}
\usepackage{times}
\usepackage{helvet}
\usepackage{courier}
\usepackage{tikz}
\usepackage{graphicx}
\usepackage{epsfig}
\usepackage{amssymb}
\usepackage{color}
\usepackage[hyphens]{url}
\usepackage{array}
\usepackage{tabu}
\usepackage{graphicx}
\usepackage{enumerate}
\usepackage{csquotes}
\usepackage{qtree}
\usepackage{relsize}
\usepackage{caption}
\usepackage{subcaption}
\usepackage{booktabs}
\usepackage{float}
\restylefloat{table}
\usepackage{rotating}
\usepackage[symbol,bottom,flushmargin,hang,multiple]{footmisc}
\usepackage{multirow}
\usepackage{hhline}
\usetikzlibrary{automata,topaths}
\usepackage{mathtools}
\usepackage{algorithm}
\usepackage{algpseudocode}
\usepackage{pifont}
\usepackage{amsmath}
\usepackage{amsfonts}
\usepackage{enumitem}
\usepackage{setspace}
\usepackage{rotating}
\usepackage{flushend}

\let\oldReturn\Return
\renewcommand{\Return}{\State\oldReturn}

\begin{document}
%
\title{Space-Time Graph Modeling of Ride Requests Based on Real-World Data}
                     
\author{Abhinav Jauhri,\textsuperscript{1} Brian Foo,\textsuperscript{2} J\'er\^ome Berclaz,\textsuperscript{2} Chih Chi Hu,\textsuperscript{1} \\ {\Large \bf Radek Grzeszczuk,\textsuperscript{2} Vasu Parameswaran,\textsuperscript{2} John Paul Shen\textsuperscript{1}}\\
\textsuperscript{1}Carnegie Mellon University, USA;    \textsuperscript{2}Uber Technologies, Inc., USA\\ 
\{ajauhri, chihhu, jpshen\}@cmu.edu;  \{bfoo, jrb, radek, vasu\}@uber.com\\
}
\maketitle
\begin{abstract}
This paper focuses on modeling ride requests and their variations over location and time, based on analyzing extensive real-world data from a  ride-sharing service. We introduce a graph model that captures the spatial and temporal variability of ride requests and the potentials for ride pooling. We discover these ride request graphs exhibit a well known property called  ``densification power law" often found in real graphs modelling human behaviors. We show the pattern of ride requests and the potential of ride pooling for a city can be characterized by the \textit{densification factor} of the ride request graphs. Previous works have shown that it is possible to automatically generate synthetic versions of these graphs that exhibit a given densification factor. We present an algorithm for automatic generation of synthetic ride request graphs that match quite well the densification factor of ride request graphs from actual ride request data.
\end{abstract}

\section{Introduction}
Recent emergence of ride-sharing services is transforming human mobility and transportation in major cities of the world~\cite{la2016}. In December 2015, Uber Technologies, Inc. reported completion of a billion rides~\cite{uber2015} within five years since it started operations. Didi alone in China reported 1.4 billion ride requests in 2015~\cite{didi2016}. There is huge potential for such services to transform urban transportation, public policies, and city-scale services. 

Prior works have assessed the potential benefits of ride sharing services. More efficient human transportation at the city scale can play a key role in contributing to sustainability. Most previous studies were based on limited amount of data from a handful of cities over a short span of time. Our work is based on extensive ride request data from the Uber ride-sharing service, which has a global footprint covering several hundred cities. We examine ride request data from 40 cities across the world covering a time span of many weeks.

\begin{figure}[!t]
  \centering
    \includegraphics[width=\linewidth]{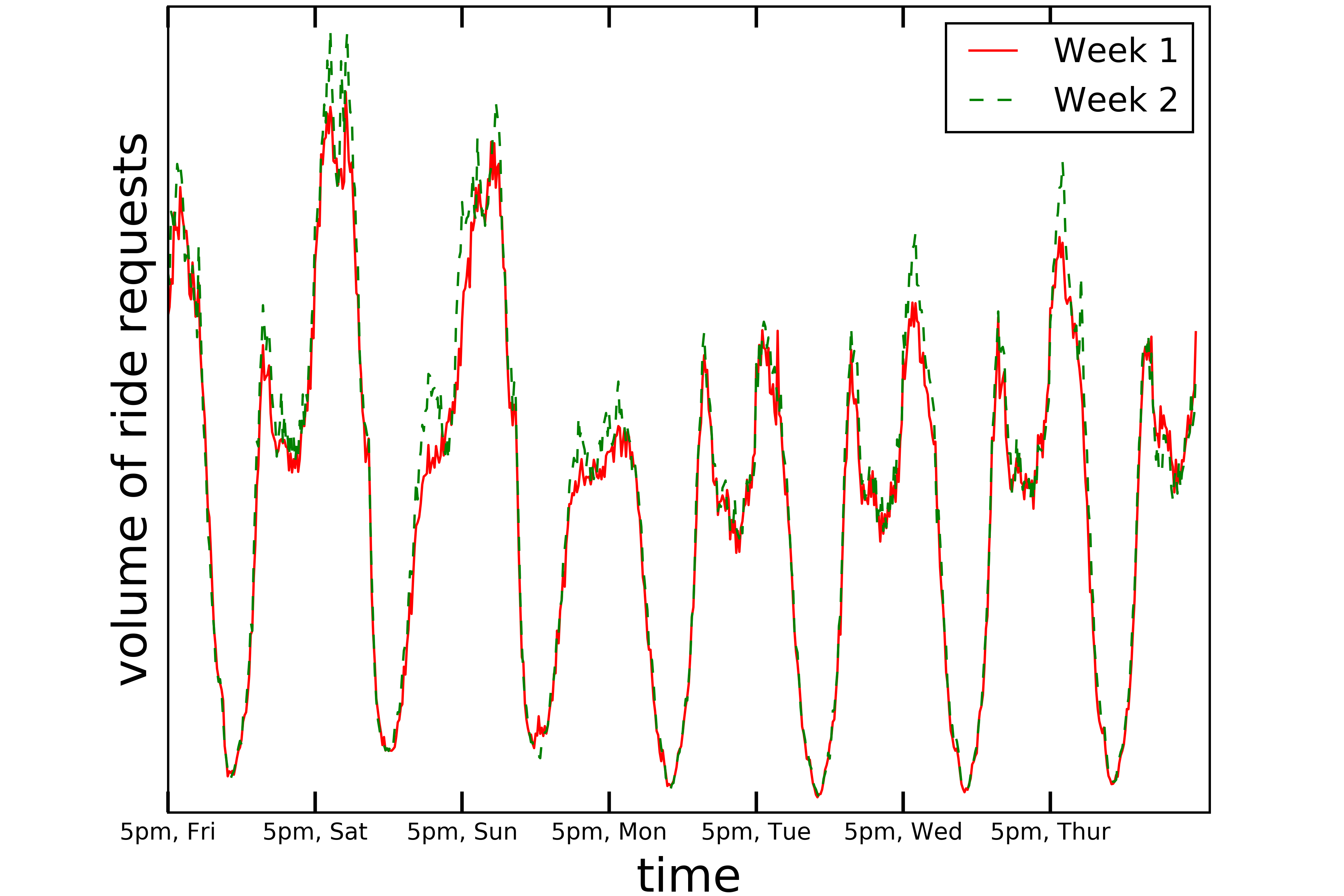}
    \caption{Similarity in the weekly pattern of ride requests for San Francisco for two different weeks.}
    \label{fig:volume_time_series}
\end{figure}

Examining the extensive ride request data from Uber, we quickly observed that the ride request patterns exhibit significant variability from city to city. Furthermore, within each city, ride requests also vary across regions of the city, on different days of the week, and at different times of the day. However, we also observed that, for most cities, there is a strong pattern that tends to repeat on a weekly basis, as shown for San Francisco in Figure~\ref{fig:volume_time_series}. Hence, effective modeling of ride requests must capture the variability in both the spatial and temporal dimensions, but can use one week as the representative time period.

Several ride-sharing services have recently introduced the notion of ride pooling (combining multiple ride requests into one vehicle) for improving overall service efficiency and at the same time reducing the number of vehicles on the road which can potentially help alleviate traffic congestion. Effective ride pooling requires bundling ride requests that occur in close proximity in both time and location.

This paper makes the following five key contributions: (1) We introduce a new graph model of ride-sharing services that captures both the temporal and spatial attributes of ride requests; (2) We discover that ride request graphs (RRGs) from this model exhibit the well known ``densification power law'' (DPL) property often found in real graphs modeling human behaviors~\cite{chakrabarti2012graph}; (3) We show it is possible to automatically generate synthetic versions of RRGs that exhibit the same DPL degree as the RRGs extracted from real world data; (4) We introduce a new concept called ``ride poolability'' that captures the fraction of ride requests that can potentially be pooled; and (5) We show there is a direct correlation between the DPL degree of RRGs and the level of ride poolability of a city.

The paper is organized as follows: In Section~\ref{sec:prior_work} we provide a survey of related work. In Section~\ref{sec:space_time_model_based_on_real_data} we present space-time evolution of ride requests based on extensive ride request data from cities around the world. We then introduce a concise space-time graph model of ride requests in Section~\ref{sec:space_time_graph_model}. We show that ride request graphs (RRG) exhibit the well known densification power law (DPL) often found in real graphs. In Section \ref{sec:synthesized_graph_based_model} we show that it is possible to automatically generate synthetic version of RRGs that exhibit the same DPL degree as RRGs extracted from actual ride requests. In Section~\ref{sec:ride_request_poolability} we introduce ride poolability and show the direct correlation between the DPL degree and the level of ride poolability. Finally, we summarize our key findings and suggest promising directions for future work in Section \ref{sec:conclusion}.

\section{Prior Work}\label{sec:prior_work}
Recent studies looked at different formulations to show the potential of ride pooling. \cite{burns2013transforming} uniformly distribute ride requests in a geographical area. They model the performance of fleet of vehicles as a queuing system where vehicles are servers and trip requests are customers. Using such an analytical model they derive average capacity utilizations, wait times, and total costs. \cite{lu2014optimization} study optimizing the number of miles driven by drivers by pooling riders; ride requests are generated uniformly over square blocks. \cite{wang2013optimizing} also simulate data for ride requests in the city of Atlanta to study benefits of matching riders. \cite{knapen2015scalability} formulate a graph with nodes as users, and edges indicating whether or not a negotiation between two users is possible to carpool. The data used for users here was a synthetic population for a geographical area. \cite{kamar2009collaboration} use real-world data on trips from a limited part of a city to highlight significant reduction of carbon dioxide per year by having multiple riders share the same vehicle. \cite{bicocchi2014investigating} developed a recommender system capable of identifying riders which could be pooled by looking at users' location data when they send or receive calls or text messages. \cite{stiglic2015benefits} study the benefits of meeting points in a ride sharing system from generating synthetic ride requests within limited distance. \cite{shmueli2015ride} use a graph model to analyze real data set of taxi trips in New York City for assessing the potential of ride pooling. 

Previous works have used either synthetic models to generate data or real data which may not accurately represent locations where ride requests originate or terminate. Even if it is representative, temporal changes which affect ride requests have not been considered. For instance, users' locations for making phone calls in the afternoon from offices may not necessarily imply people travel often at the same time from offices. In fact, comparatively low number of ride requests are associated with afternoons on any weekday. For ride pooling to be some percentage of total ride requests such that civic bodies can make decisions, or for ride requests pattern to be studied for its potential of pooling~\cite{huang2014large} and to improve arrival time~\cite{cao2016multiagent} based on anticipated congestion, we need to understand the laws which govern ride requests at any geographical area, and at any given time. 

This work models ride requests as a graph. Some important prior works on graphs have helped us to model and discover properties for ride requests. There is a class of graphs that models real networks, e.g. social network graphs and publication citation graphs, that evolve over time. These graphs become more and more dense as they evolve in time, i.e. the edge count grows superlinearly relative to the node count growth. This \textit{densification} of the graph can be modeled concisely by a power law relationship between the edge count and the node count. Graphs for many real networks all seem to exhibit this densification power law (DPL)~\cite{newman2005power}. For this class of graphs it is possible to automatically generate synthetic graphs that exhibit the same densification power law without needing the original data~\cite{leskovec2005graphs}. These graphs also exhibit the attribute of having strongly connected subgraphs or communities~\cite{fortunato2010community}. In this work, we have discovered that our space-time graph model of ride requests for a city belongs to this class of graphs. This fact allows us to discover interesting attributes and insights about ride requests and the potential of ride pooling for ride-sharing services.

\section{Space-Time Evolution of Ride Requests}\label{sec:space_time_model_based_on_real_data}
This work is based on extensive real world data from Uber, a ride-sharing service with global presence in several hundred cities. We examine data from 40 cities with average daily city-wide ride requests ranging from 2K to 200K per day.
\footnote{This paper presents results for only four cities. We have similar results for all 40 cities based on a total of about 50M ride requests.} 
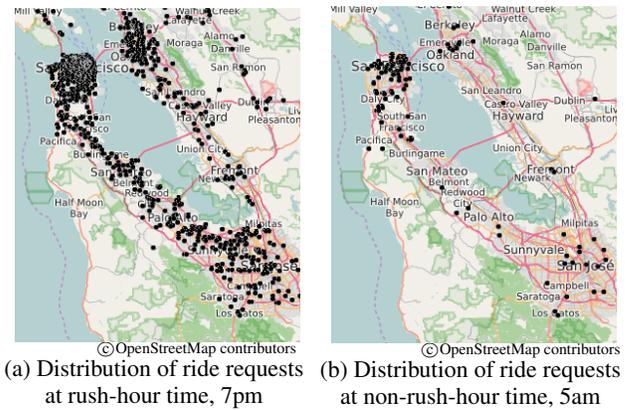
\begin{figure}[!t]
\captionsetup{justification=centering}
\centering
\begin{subfigure}[!t]{.5\linewidth}
  \centering
   \begin{tikzpicture}[remember picture, overlay]
    \node[draw=none,align=right] at (2.5,-0.1) {\tiny \copyright OpenStreetMap contributors};
  \end{tikzpicture}
    \includegraphics[width=0.9\linewidth]{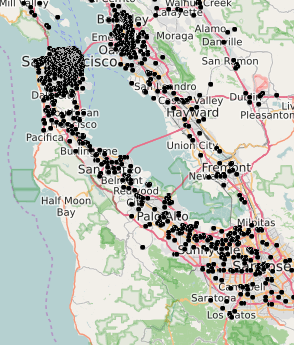}
    \caption{Distribution of ride requests at rush-hour time, 7pm}
    \label{fig:san_francisco_0}
\end{subfigure}%
\begin{subfigure}[!t]{.5\linewidth}
  \centering
   \begin{tikzpicture}[remember picture, overlay]
    \node[draw=none,align=right] at (2.6,-0.1) {\tiny \copyright OpenStreetMap contributors};
  \end{tikzpicture}
    \includegraphics[width=0.9\linewidth]{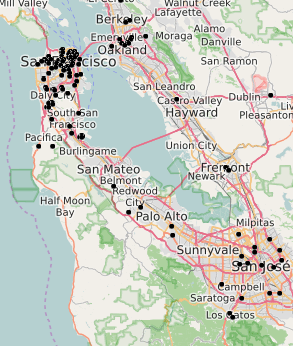}
    \caption{Distribution of ride requests at non-rush-hour time, 5am}
    \label{fig:san_francisco_120}
\end{subfigure}
\caption{Space-time variability: Each dot represents a source or destination for a ride request in San Francisco\protect\footnotemark}
\label{fig:spatial_temporal}
\end{figure}
\footnotetext{Map data of Figures~2, 3, and 5  are available  from OpenStreetMap under the Open Database License and the cartography is licensed under the Creative Commons Attribution-ShareAlike 2.0. \url{http://www.openstreetmap.org/copyright}}

Each ride request involves a source location $s$, destination location $d$, and the time of the request $t$. Both $s$ and $d$ are represented by their latitude and longitude. Each ride request is considered independent of any other ride request. In this study we do not consider the actual navigation path taken from $s$ to $d$ for a ride request.

Based on examining the ride request data from the 40 cities gathered over several months in the Spring of 2016, we can make two key observations. For each city, the spatial and temporal ride request patterns tend to repeat from week to week. On the other hand, there is significant variability of ride request patterns from city to city. Furthermore, in each city there is variability across different days of the week, at different times of the day, and across different regions of the city. 

Variability in ride request patterns in San Francisco for two snapshots taken at two different times of the day is shown in Figure~\ref{fig:spatial_temporal}. The figure shows the spatial distribution of ride request density over the Bay Area. San Francisco downtown (top left cluster of points) is clearly denser in ride requests. The two figures illustrate two snapshots of ride request density for two different 5-minute intervals one at 7:00pm and the other at 5:00am. The temporal and spatial variations of ride requests can be clearly seen.

\section{Space-Time Graph Model}\label{sec:space_time_graph_model}
We now introduce a graph model for ride requests. The graph representing ride requests for a specific time period is called a \textit{Ride Request Graph} (RRG). Each RRG captures the spatial distribution of ride requests across a city within that time period. The RRG evolves from one time period to the next, to account for new ride requests initiated in the next time period. This evolution of the RRG and the resultant sequence of RRGs capture the temporal aspect of ride requests over many time intervals.

\subsection{Ride Request Graph Generation}

For a given time interval we can generate a RRG representing all the ride requests in that interval. We divide the map of a city into equal sized cells of $100m\times100m$ each. Each cell is considered as a node in the graph only if the source or destination of a ride request falls within that cell. A directed edge connects the source and destination cells of a ride. A directed graph can then be generated to model all the ride requests in that time interval for a given city. 

\begin{figure}[!t]
\captionsetup{justification=centering}
\centering
\begin{subfigure}[t]{.5\linewidth}
  \centering
  \includegraphics[width=4cm,height=4cm]{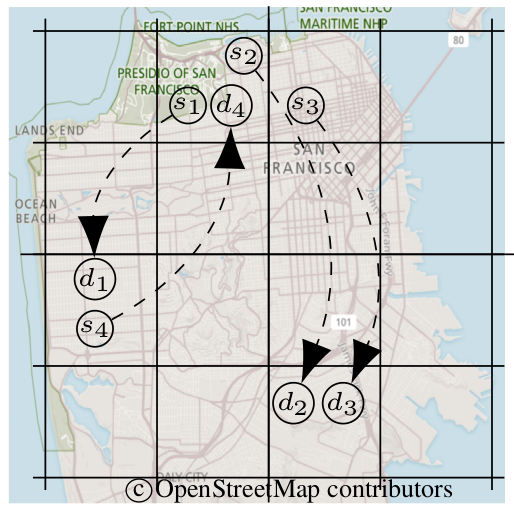}
  \caption{Four ride requests distributed spatially over a map}
\label{fig:source_dest_points}
\end{subfigure}%
\hfill
\begin{subfigure}[t]{.5\linewidth}
  \centering
    \includegraphics[width=4cm,height=4cm]{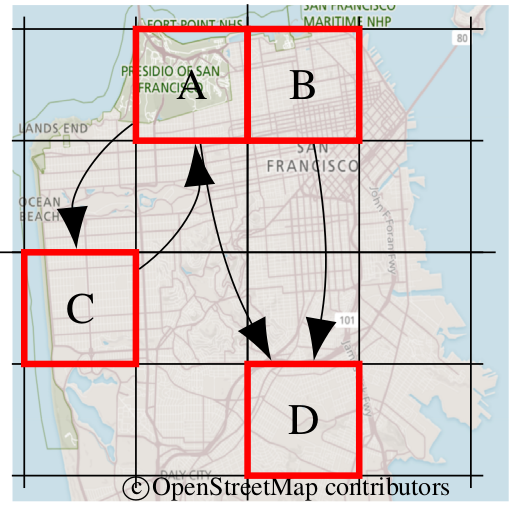}
    \caption{Corresponding Ride Request Graph with four nodes (marked by red boxes) and directed edges.}
\label{fig:graph_from_points}
\end{subfigure}
\caption{Transformation of ride requests, in a particular time interval, into a directed ride-request graph (RRG).}
\label{fig:graph_formulation}
\end{figure}

For illustration, consider the ride requests in Figure~\ref{fig:source_dest_points} for a given time interval. The four ride requests are shown on a gridded map (not drawn to scale). The corresponding graph in Figure \ref{fig:graph_from_points} is formed by four nodes with node $A$ subsuming $s_1, s_2, d_4$; node $B$ subsuming $s_3$; node $C$ subsuming $s_4, d_1$; and node $D$ subsuming $d_2, d_3$. All edges in Figure \ref{fig:source_dest_points} have unit weights representing single ride requests. Edge weights represent the number of ride requests from the same source and destination nodes. In this paper, the term \textit{node} is always used in the context of the RRG graph. We use the term \textit{point} to refer to a specific location defined by its latitude and longitude, which could be the source or destination of a ride request, and is associated with a node of RRG.

\subsection{Ride Request Graph Densification}

\begin{figure*}[!t]
\captionsetup{justification=centering}
\centering
\begin{minipage}[t]{\linewidth}
\begin{subfigure}{.25\textwidth}
  \centering
    \includegraphics[width=\linewidth, height=5cm]{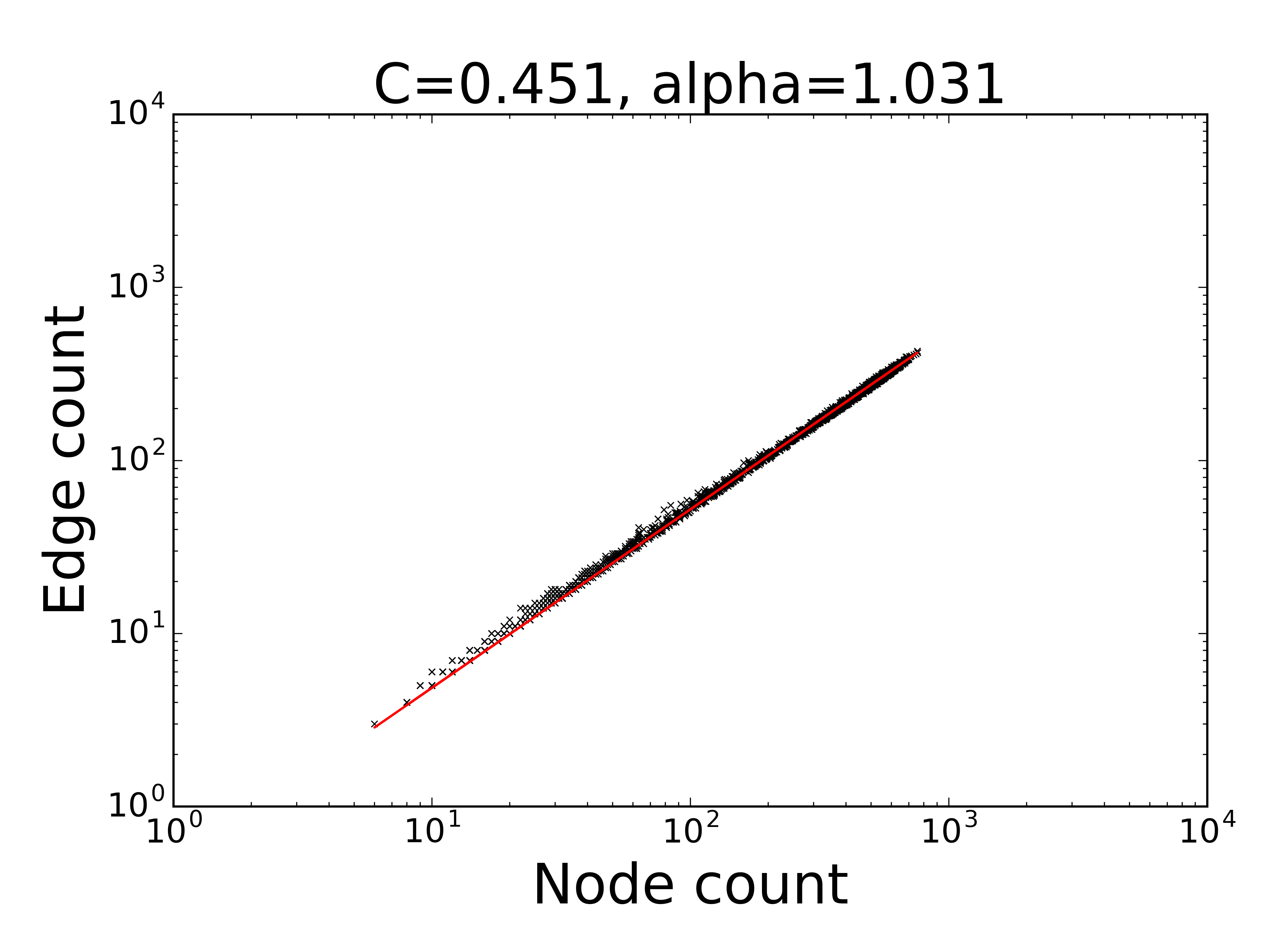}
    \label{fig:real_hyderabad_edge_node}
\end{subfigure}%
\begin{subfigure}{.25\textwidth}
  \centering
    \includegraphics[width=\linewidth, height=5cm]{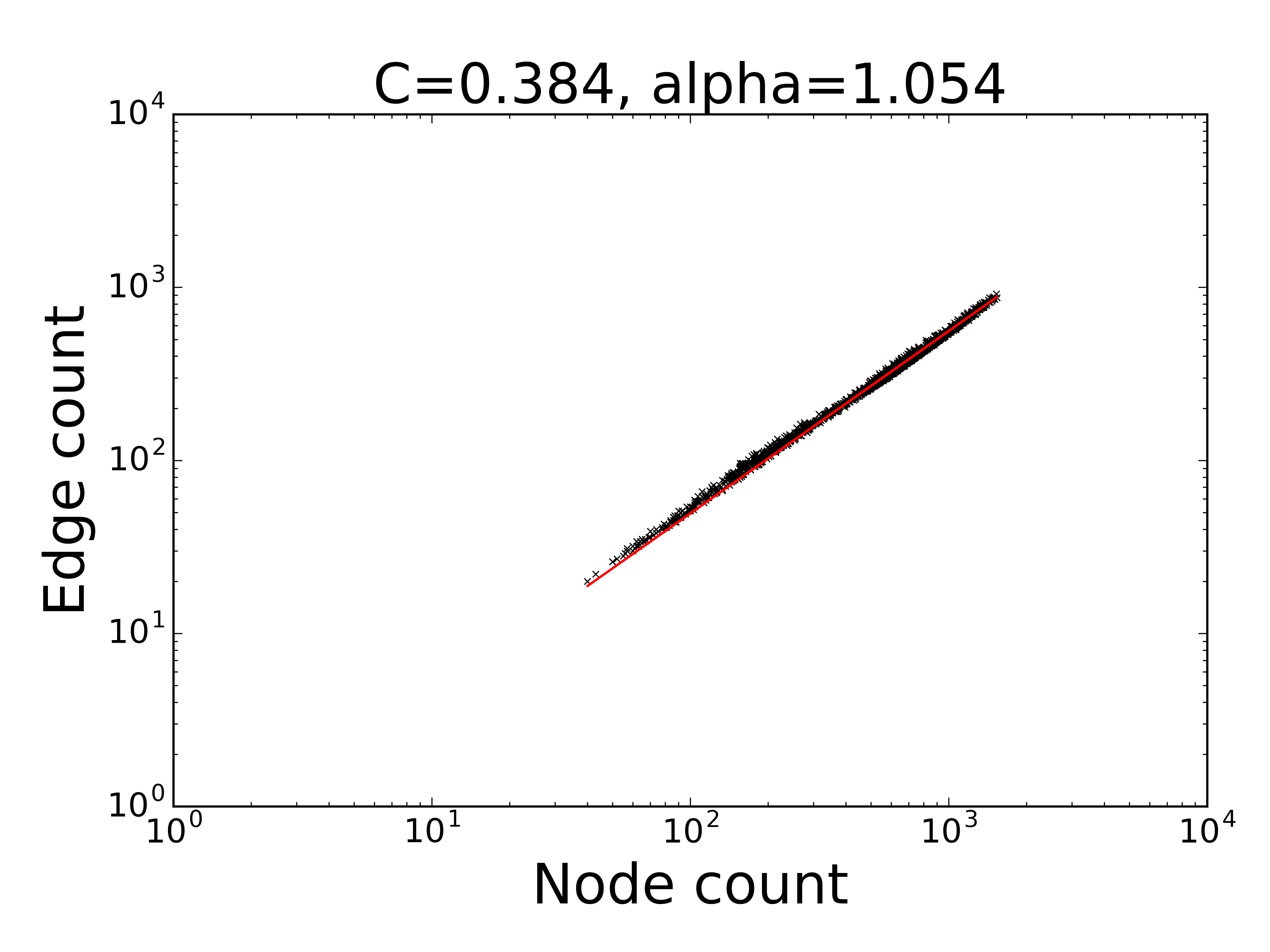}
    \label{fig:real_paris_edge_node}
\end{subfigure}%
\begin{subfigure}{.25\textwidth}
  \centering
    \includegraphics[width=\linewidth, height=5cm]{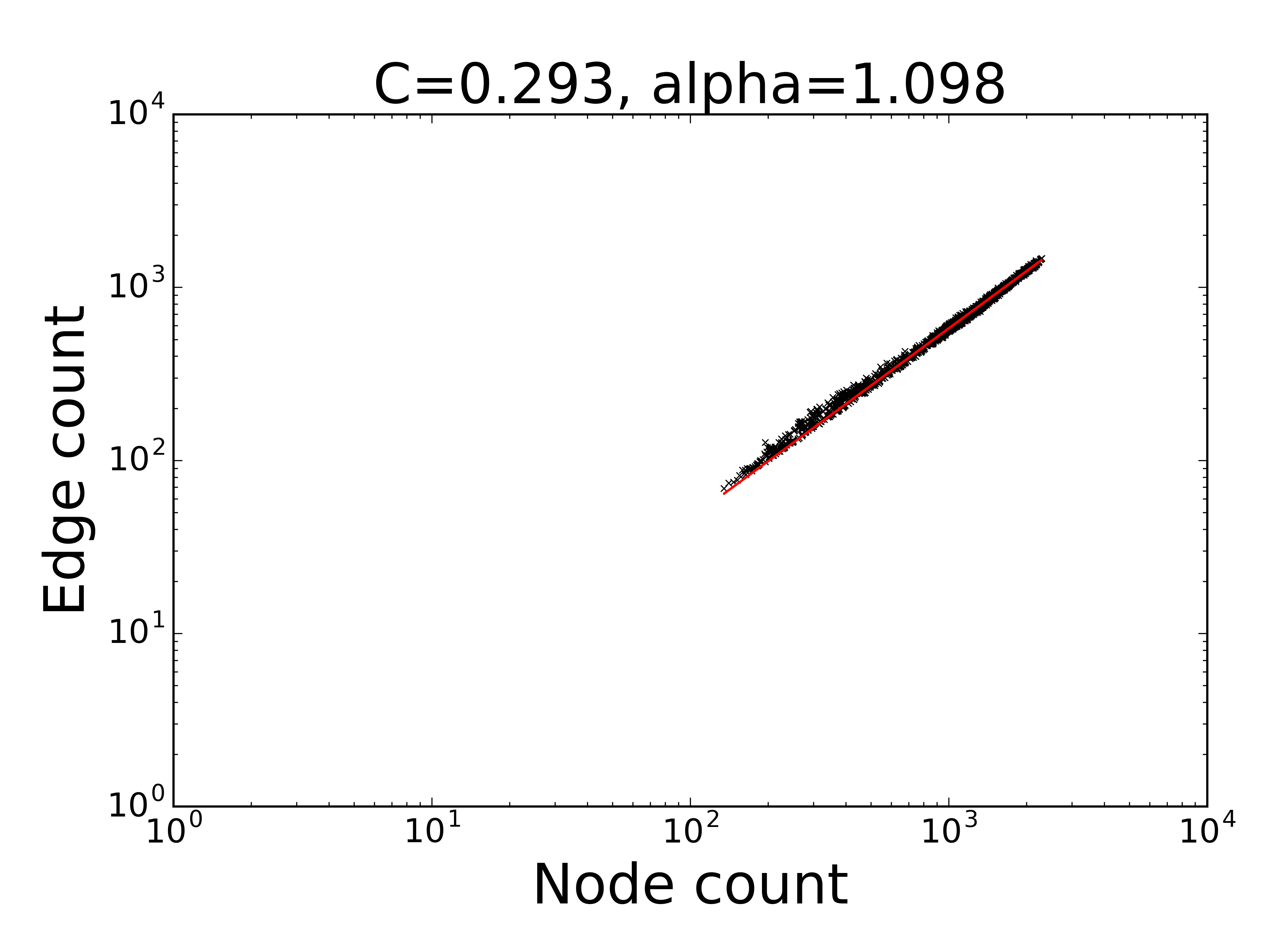}
    \label{fig:real_new_york_edge_node}
\end{subfigure}%
\begin{subfigure}{.25\textwidth}
  \centering
    \includegraphics[width=\linewidth, height=5cm]{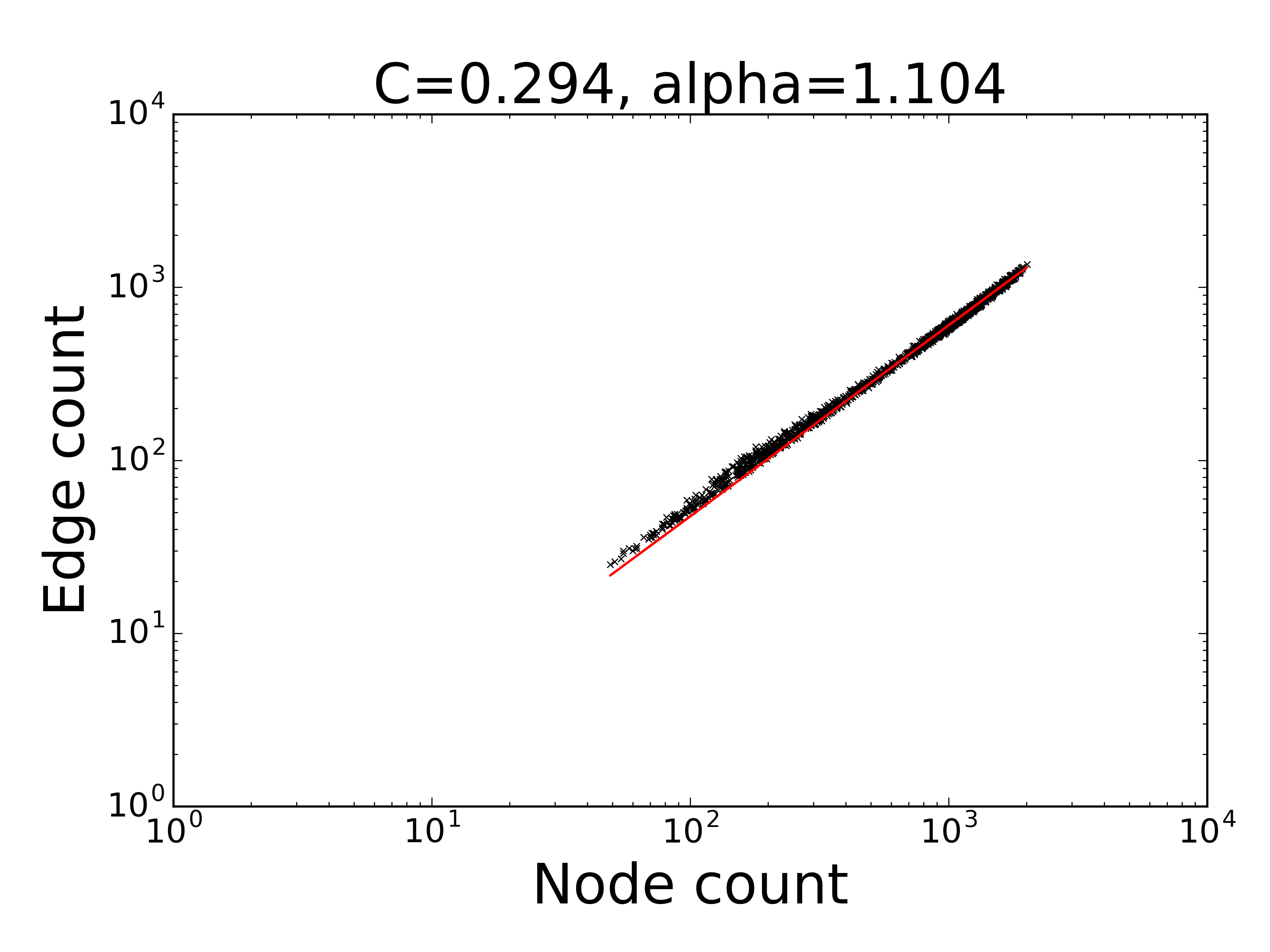}
    \label{fig:real_san_francisco_edge_node}
\end{subfigure}
\end{minipage}\\[-4ex]
\begin{minipage}[t]{\linewidth}
\begin{subfigure}{.25\textwidth}
  \centering
    \includegraphics[width=\linewidth, height=5cm]{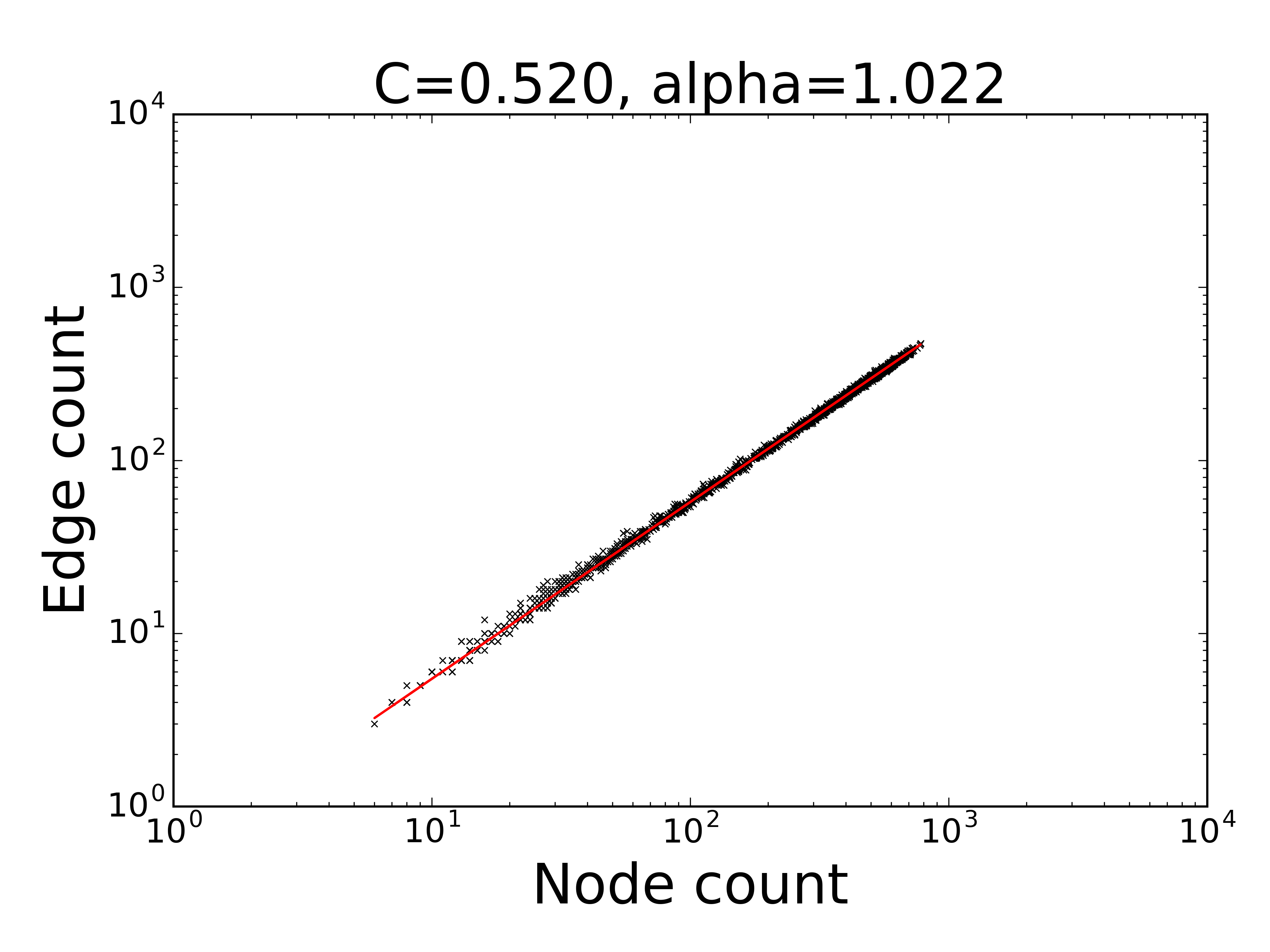}
    \caption*{Hyderabad}
    \label{fig:rndm_hyderabad_edge_node}
\end{subfigure}%
\begin{subfigure}{.25\textwidth}
  \centering
    \includegraphics[width=\linewidth, height=5cm]{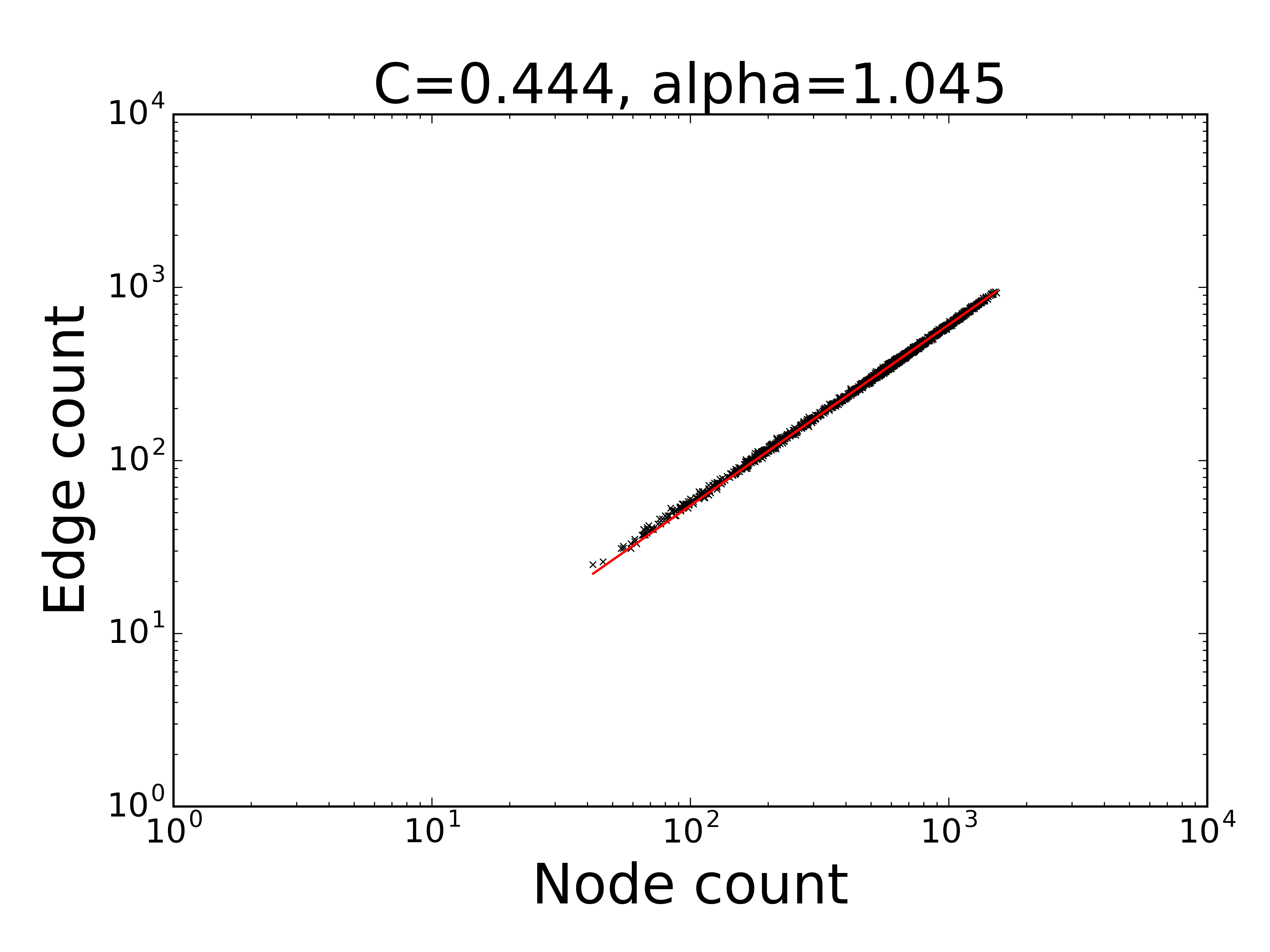}
    \caption*{Paris}
    \label{fig:rndml_paris_edge_node}
\end{subfigure}%
\begin{subfigure}{.25\textwidth}
  \centering
    \includegraphics[width=\linewidth, height=5cm]{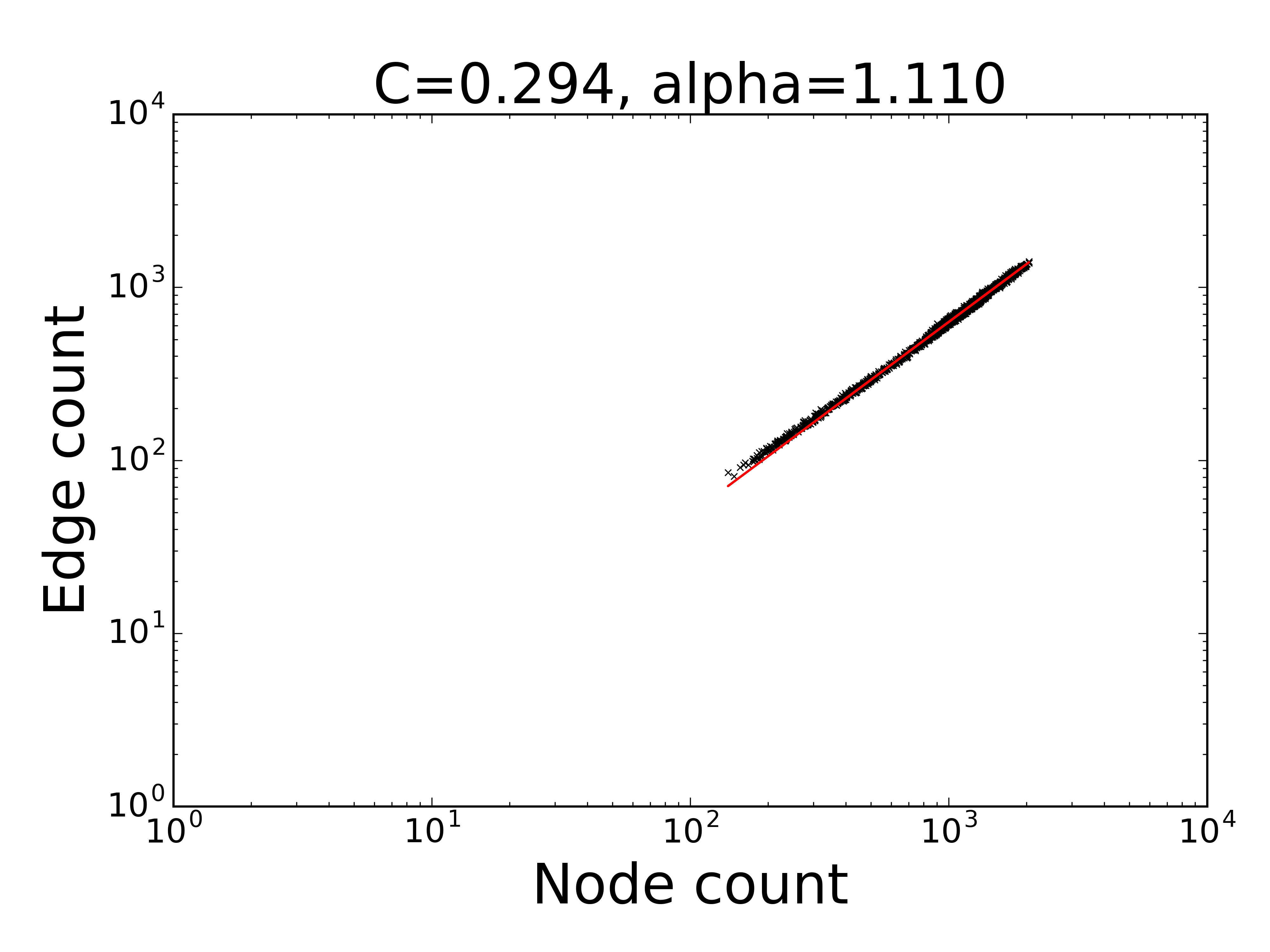}
    \caption*{New York}
    \label{fig:rndm_new_york_edge_node}
\end{subfigure}%
\begin{subfigure}{.25\textwidth}
  \centering
    \includegraphics[width=\linewidth, height=5cm]{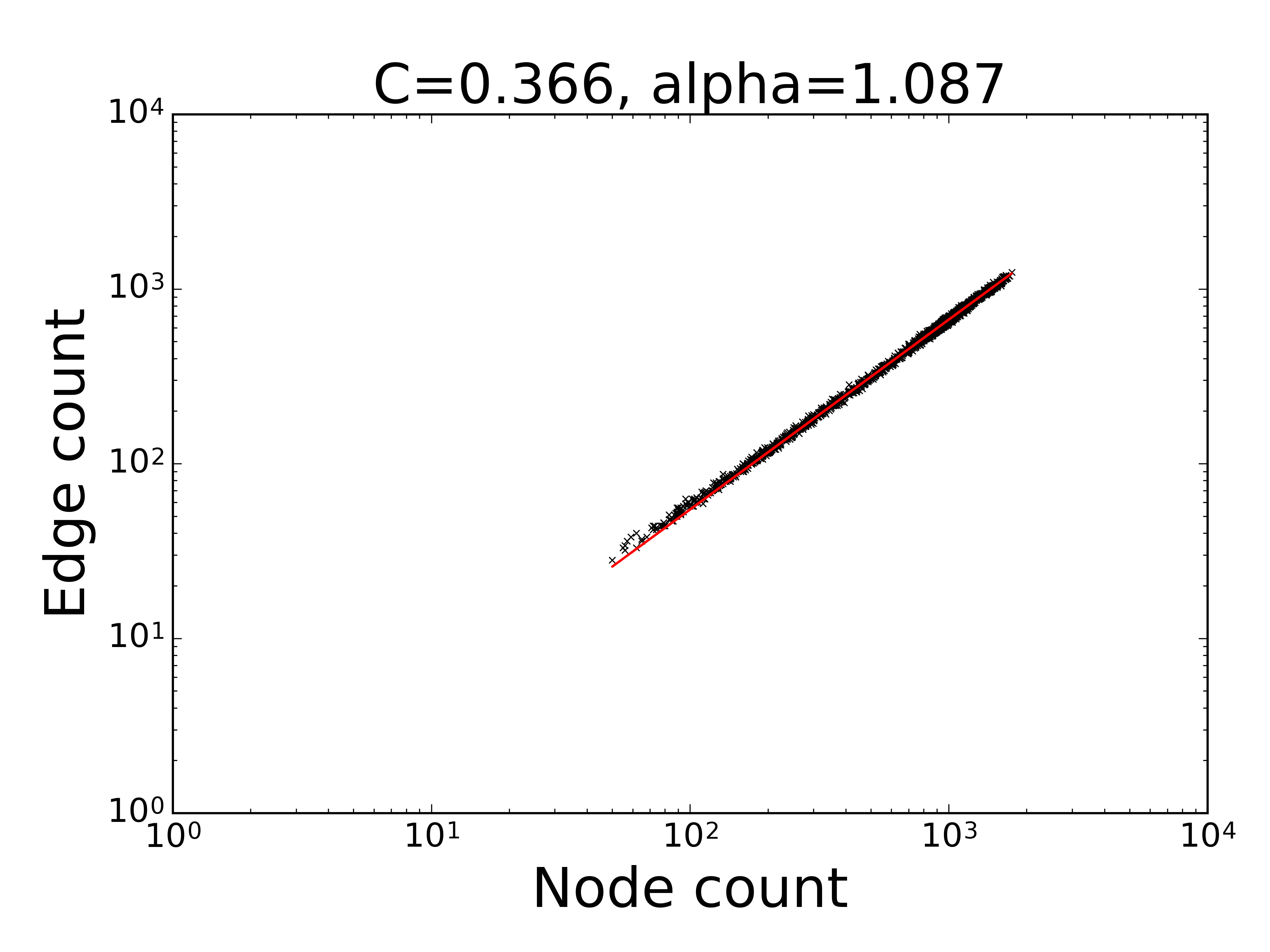}
    \caption*{San Francisco}
    \label{fig:rndm_san_francisco_edge_node}
\end{subfigure}
\end{minipage}
\caption{DPL plots from real data (top row) and synthetic data (bottom row) for four cities. The red line is the least square fit of the form $y = Cx^{\alpha}$, where $y$ and $x$ are number of edges and nodes respectively. $R^2 \approx 1.00$ for all of them.}
\label{fig:edge_node}
\end{figure*}

A Ride Request Graph is different from conventional graphs: (1) each node has a geographical area associated with it; (2) RRG is not fixed in time but evolves in time. Each RRG involves a spatial quantization (into $100m\times100m$ cells) of the geographical space of a city, and a temporal quantization into sequence of time intervals. In this work we use 5-minute intervals for temporal quantization. As an RRG evolves in time, it produces a sequence of RRGs that capture the temporal behavior of ride requests.

As we analyze the RRGs extracted from historical data of ride requests from all the cities, we make an interesting observation about these RRGs. \textit{Densification} refers to graphs that evolve in time, and how the edge count grows relative to the growth of the node count. Many graphs modeling aspects of human behaviors, such as social network graphs and publication citation graphs, among others, exhibit densification over time that follows a power law, i.e. the number of edges grows as a power of the number of nodes~\cite{leskovec2007graph}. We have discovered that the RRGs for ride requests exhibit the same power law densification behavior and belongs to this class of graphs. 

We observe RRGs at different snapshots of time, with each spanning five minutes. For each snapshot, we study the \textit{Densification Power Law} plot (DPL plot)~\cite{leskovec2005graphs} i.e. log-log plot of the number of edges $e(t)$ versus number of nodes $n(t)$.

Top row in Figure~\ref{fig:edge_node} shows the Densification Power Law (DPL) plot for four cities based on real data for a typical week in 2016. It is observed that for every time interval $t$:
\begin{equation}
\begin{split}
e(t) &\propto n(t)^\alpha \\
&= C n(t)^\alpha,
\end{split}
\label{eq:power_law}
\end{equation}
where $e(t)$ and $n(t)$ are the number of edges and number of nodes respectively, formed by all ride requests occurring in the time interval $t$. $C$ and $\alpha$ are constants. Number of edges is a good approximation of the number of ride requests. We observe that all the cities follow the densification power law but the parameters of the power law vary from city to city.
 
\subsection{Characteristic Attributes of RRGs}
It appears that the pattern of ride requests for a city can be characterized concisely by $C$, and $\alpha$ derived from the power law of RRGs for that city (see top row of Figure~\ref{fig:edge_node}). The exponential $\alpha$ depicts the densification of ride requests within a city. This \textit{densification factor} $\alpha$ can range in value from 1.0 to 2.0. If $\alpha = 1.0$, this means the number of edges is growing linearly with respect to the number of nodes; if $\alpha = 2.0$, then the RRGs become fully connected graphs. 

It is interesting to note that all four cities exhibit densification factors greater than 1.0. This means the edge count is growing superlinearly to the node count, implying the densification of ride requests. We speculate this demonstrates the human tendency towards the creation of clustered/connected communities, perhaps reflecting the \emph{small world} effect~\cite{watts1998collective}. 

DPL graphs exhibit a fascinating attribute. \cite{leskovec2005graphs} and \cite{chakrabarti2012graph} have shown that for graphs that evolve according to the densification power law, it is possible to automatically generate these graphs that exhibit specific densification factors. This means that we can automatically generate RRGs that exhibit similar densification factor as that of RRGs extracted from real data. This can potentially allow us to generate synthetic RRGs, exhibiting similar densification factor, for a city without needing the real ride request data.

\section{Synthesized Space-Time Graph Model}\label{sec:synthesized_graph_based_model}

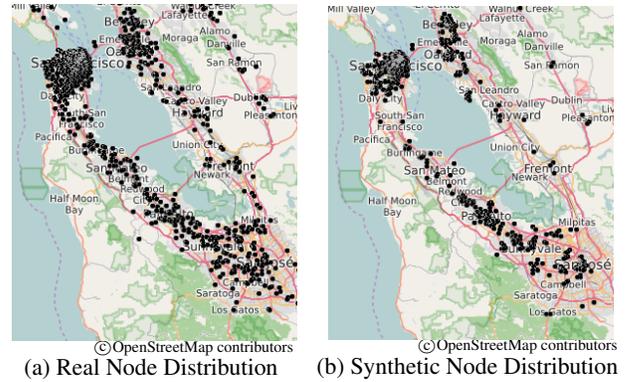
\begin{figure}[!t]
\captionsetup{justification=centering}
\centering
\begin{subfigure}{.5\linewidth}
  \centering
       \begin{tikzpicture}[remember picture, overlay]
    \node[draw=none,align=right] at (2.5,-0.1) {\tiny \copyright OpenStreetMap contributors};
  \end{tikzpicture}
    \includegraphics[width=.9\linewidth]{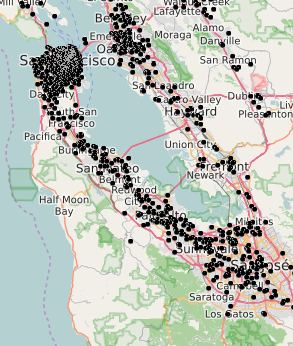}
    \caption{Real Node Distribution}
    \label{fig:san_francisco_2_real}
\end{subfigure}%
\begin{subfigure}{.5\linewidth}
  \centering
      \begin{tikzpicture}[remember picture, overlay]
    \node[draw=none,align=right] at (2.6,-0.1) {\tiny \copyright OpenStreetMap contributors};
  \end{tikzpicture}
    \includegraphics[width=.9\linewidth]{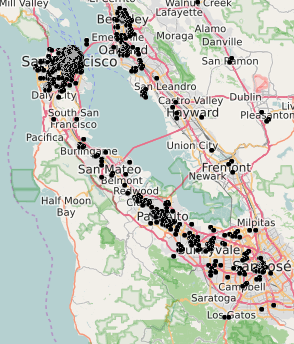}
    \caption{Synthetic Node Distribution}
    \label{fig:san_francisco_2_synthetic}
\end{subfigure}
\caption{Plots of nodes for San Francisco for a single time interval of five minutes. On the left, is the real spatial distribution, and on the right is the synthetically generated.}
\label{fig:real_and_random_nodes}
\end{figure}

In this section, we explore the automatic generation of synthesized space-time graph models that mimic the attributes of RRGs generated using real ride request data. The two attributes of interest are: 1) the spatial distribution of nodes; and 2) the temporal evolution or the densification factor of the RRG. 

\subsection{Spatial Properties}
Spatial properties provide information on the source and destination locations of ride requests as shown in Figure~\ref{fig:source_dest_points}. The synthesized graph model should capture the spatial distribution of these ride request locations.

To capture spatial properties, we compute the likelihood of a node to either possess a source or destination location by using geospatial information from OSM's public data~\cite{osm} on node density\footnote{OSM data is publicly available at \url{http://download.geofabrik.de/}}. To avoid confusion with a node of the Ride Request Graph, we shall refer to an OSM node\footnote{\url{http://wiki.openstreetmap.org/wiki/Node}} as \textit{Point of Interest} (PoI). A PoI is defined by a tuple of latitude and longitude. PoIs are used to define standalone features such as traffic signals, businesses, schools, hospitals, and many others. Alternatively, any other dataset providing a quantitative measure over geographical space which is correlated with the likelihood of ride requests can also be used.

\begin{figure}
 \captionsetup{justification=centering}
   \centering
   \begin{tikzpicture}
   [
           box/.style={rectangle,draw=red,thick, minimum size=.9cm},
           lbox/.style={rectangle,draw=red,thick, minimum size=.9cm, opacity=.2},
   ]
     \draw (-2.,0) -- (2.,0);
     \draw (0,-1.) -- (0,2.);
     \node[box] (init) at (-0.44,1.35) {\tiny $s$};
    \node[box] (rs1) at (-0.45,0.45) {\tiny $rs_1$};
     \node[box] (rs2) at (-1.35,0.45) {\tiny $rs_2$};
     \node[box] (rs3) at (-1.35,-0.45) {\tiny $rs_3$};
     \draw [dashed,opacity=.5] (init.center) to node {} (rs1.center);
     \draw [dashed,opacity=.5] (rs1.center) to node {} (rs2.center);
     \draw [dashed,opacity=.5] (rs2.center) to node {} (rs3.center);
     \draw[step=.9cm] (-1.9,-1) grid (1.9,1.9);

     \node[lbox,fill=blue] at (-1.35,-0.45){};
 \end{tikzpicture}
 \caption{Random walk starting at node $s$ with three random steps to reach node $rs_3$. Final point is selected uniformly random within the blue colored grid.}
 \label{fig:random_walk}
\end{figure}
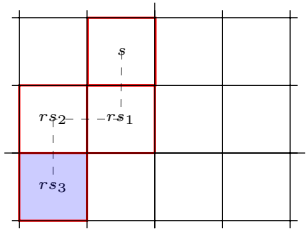

Algorithm~\ref{alg:spatial_prop_gen} performs node selection using vector of probabilities $pr \in \mathbb R^n, n=\left\vert{S}\right\vert$, where $S$ is a subset of nodes of interest. $pr$ is computed by aggregating all PoIs present at a RRG node, and then normalizing to get the probability mass function across nodes in $S$. Algorithm~\ref{alg:spatial_prop_gen} takes as input the number of synthetic points $m$ to be generated, and associates each synthetic point to an initial node; this is determined from prior probability vector $pr$. Once the initial node is chosen, Algorithm~\ref{alg:random_walk} performs a random walk starting from initial node centroid such that the synthetic points are spread out in the geographical area (Figure~\ref{fig:random_walk}). Since PoI data could be sparse, we use a kernel density estimation function $K$\footnote{We used Gaussian Kernel Density Estimation library \url{http://statsmodels.sourceforge.net/devel/generated/statsmodels.nonparametric.kernel_density.KDEMultivariate.html}} over the geographical space to guide the random walk. 

In Algorithm~\ref{alg:spatial_prop_gen} method \emph{randomChoice} generates $m$ points with replacement using the prior probabilities vector $pr$; \emph{geoCoords} returns the latitude and longitude associated with the node label; \emph{perturb} performs a uniform random selection within the final node (blue area in Figure~\ref{fig:random_walk}) to determine the final location of the newly generated point.

In Algorithm~\ref{alg:random_walk}, method \emph{randomStep} chooses a node amongst the neighbouring eight nodes (or less) of the current node, $curr$, by normalizing the probabilities returned by the kernel density estimates; it returns the new node $new$, and a reward $r$. We kept reward equal to the probability estimate for current node, $r = K(curr)$. Note that every node is defined by latitude, longitude coordinates of its centroid. 

\begin{algorithm}[!t]
\begin{spacing}{0.8}
\scriptsize
\begin{algorithmic}[1]
\Procedure{spatialPropGen}{$K, m, pr$}
\State $labels=$ \emph{randomChoice}$([0, \ldots, \text{len}(pr)], m, pr)$
\State $pts = []$
\For{each point $i$ \Pisymbol{psy}{206} $[0, m)$ }
\State $l = labels[i]$ 
\State $s=$ \emph{geoCoords}$(l)$
\State $p =$ \textsc{randomWalk}{$(K, s, max_r, max_{s})$}
\State add$(pts$, \emph{perturb}$(p))$
\EndFor
\Return $pts$
\EndProcedure
\end{algorithmic}
\caption{To capture spatial properties. Inputs: Kernel density estimation function $K$, number of synthetic points $m\in \mathbb N$, prior probability vector $pr \in \mathbb R^n$ }
\label{alg:spatial_prop_gen}
\end{spacing}
\end{algorithm}

\begin{algorithm}[!t]
\begin{spacing}{0.8}
\scriptsize
\begin{algorithmic}[1]
\Procedure{randomWalk}{$K, s, max_r, max_{s}$}
\State let $tot_r$ = 0
\State let $n_{steps}$ = 0
\State let $curr = s$
\While{$tot_r \leq max_r$ and  $n_{steps} \leq max_{s}$}
\State $curr~,~r =$ \emph{randomStep} $(curr, K)$
\State $n_{steps} = n_{steps} + 1$
\State $tot_r = tot_r + r$
\EndWhile
\Return $curr$
\EndProcedure
\end{algorithmic}
\caption{Random Walk. Inputs: Kernel density estimation function $K$, start location $s$, maximum reward, and maximum number of steps $max_r~\&~max_s$}
\label{alg:random_walk}
\end{spacing}
\end{algorithm}

\subsection{Densification Properties}
In the previous subsection, we only distribute points spatially such that they are either source or destination locations. Our model to connect these points such that they become concrete ride requests is described in Algorithm~\textsc{densPropGen}. This model allows us to capture the \textit{densification} property observed in RRGs from real data. Algorithm~\ref{alg:temporal_prop_gen} requires three parameters: (1) number of points to generate $m$; (2) the probability of choosing a point which has not been visited before $p$; (3) number of outlinks $n_{edges}$ from a source point is defined by geometrically distributed random number with mean $1/q$. $1-p$ is the probability of choosing a previously visited source point as destination (variation of the preferential attachment technique described in~\cite{chakrabarti2012graph}) which captures the idea of \textit{rich getting richer}.

\begin{algorithm}
\begin{spacing}{0.8}
\scriptsize
\begin{algorithmic}[1]
\Procedure{densPropGen}{m,p,q}
\State let $M = [0, \ldots, m-1]$
\State let $R = []$
\State let $rides = []$
\While{length(M) $> 1$}
\State $s$ = \emph{uniformRandomChoice}$(M)$
\State remove$(M, s)$
\State $n_{edges}$  = \emph{geometricRandom}$(q)$
\For{each edge $e$ \Pisymbol{psy}{206} $[0, n_{edges})$ }
\If{\emph{uniformRandom}$() < p$}
\State $d$ = \emph{uniformRandomChoice}$(M)$
\State remove$(M, d)$
\Else
\State $d$ = \emph{uniformRandomChoice}$(R)$
\EndIf
\State $r = $ connect$(s, d)$
\State add$(rides,  r)$
\EndFor
\State add$(R, s)$
\EndWhile
\Return $rides$
\EndProcedure
\end{algorithmic}
\caption{To capture densification properties. Inputs: $m \in \mathbb N$ , $p~\&~q \in [0,1]$}
\label{alg:temporal_prop_gen}
\end{spacing}
\end{algorithm}

In Algorithm~\ref{alg:temporal_prop_gen}, \emph{uniformRandomChoice} uniformly at random selects a point from the set points. \emph{geometricRandom} generates values from a geometrically distributed random variable with success probability $q$. \emph{uniformRandom} generates a uniformly random value $\in [0,1)$.



\subsection{Synthesized RRG Model}
Our complete model which embodies all aspects of the synthesized RRGs is as follows:

\begin{enumerate}[label=Step \arabic*:,itemindent=*]
\itemsep0em 
\item Use OSM data to aggregate PoI count for each node.
\item Choose a subset of nodes of interest $S$~\footnote{For our experiments we selected the nodes from historical data where ride requests happened.}.
\item Calculate prior probabilities based on PoI count to get vector $pr$ on the subset of nodes.
\item Compute Gaussian kernel density function $K$ using centroids of nodes in $S$.
\item Use \textsc{spatialPropGen} with prior probabilities, number of points to be generated, and the kernel function $K$ to generate synthetic points over space.
\item Use synthetic points to generate synthetic ride requests using \textsc{densPropGen} with parameters $p$ and $q$.
\item Create the Ride Request Graph using the synthetic ride requests returned by \textsc{densPropGen}.
\end{enumerate}
One can repeat the steps above for consecutive time intervals.

\begin{figure}[t]
\captionsetup{justification=centering}
\centering
\begin{subfigure}[t]{.5\linewidth}
  \centering
   \includegraphics[width=\linewidth, height=3.5cm]{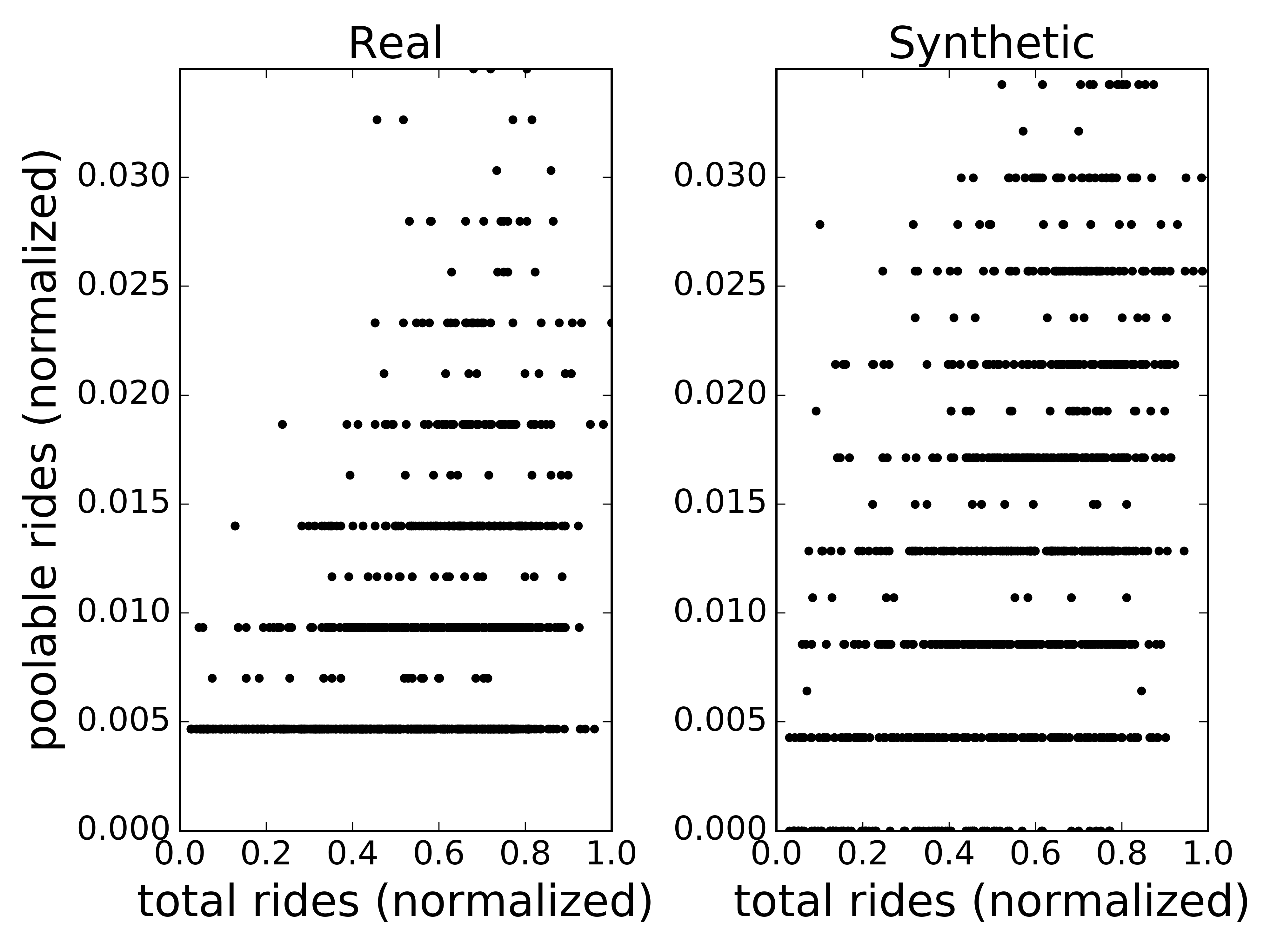}
    \caption*{Hyderabad\newline Real LS fit: slope=$0.01$ \newline Synthetic LS fit: slope$=0.018$}
    \label{fig:hyderabad_poolability}
\end{subfigure}%
\begin{subfigure}[t]{.5\linewidth}
  \centering
   \includegraphics[width=\linewidth, height=3.5cm]{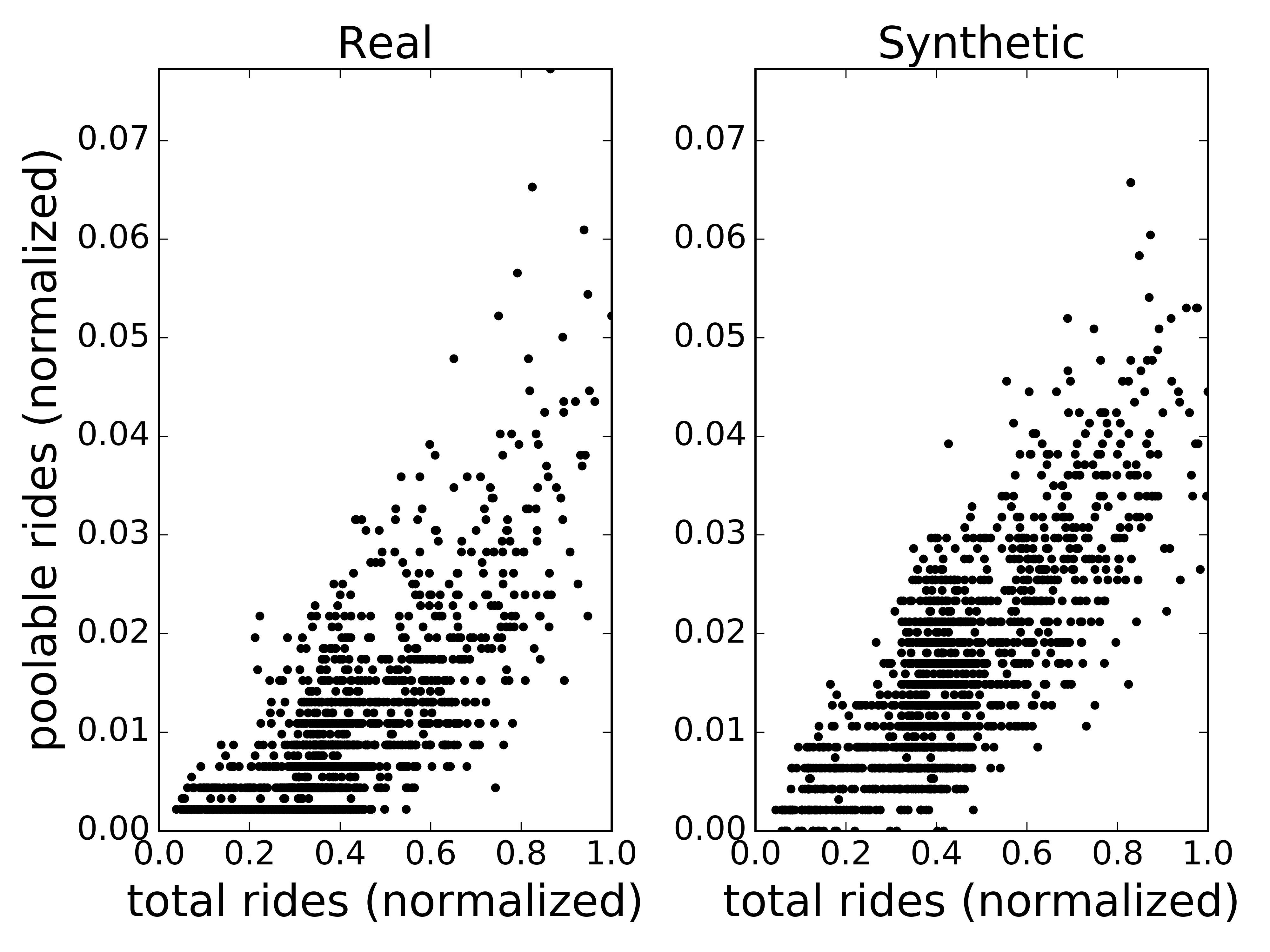}
    \caption*{Paris \newline Real LS fit: slope=$0.038$ \newline Synthetic LS fit: slope$=0.046$}
    \label{fig:paris_poolability}
\end{subfigure}

\begin{subfigure}[t]{.5\linewidth}
  \centering
  \includegraphics[width=\linewidth, height=3.5cm]{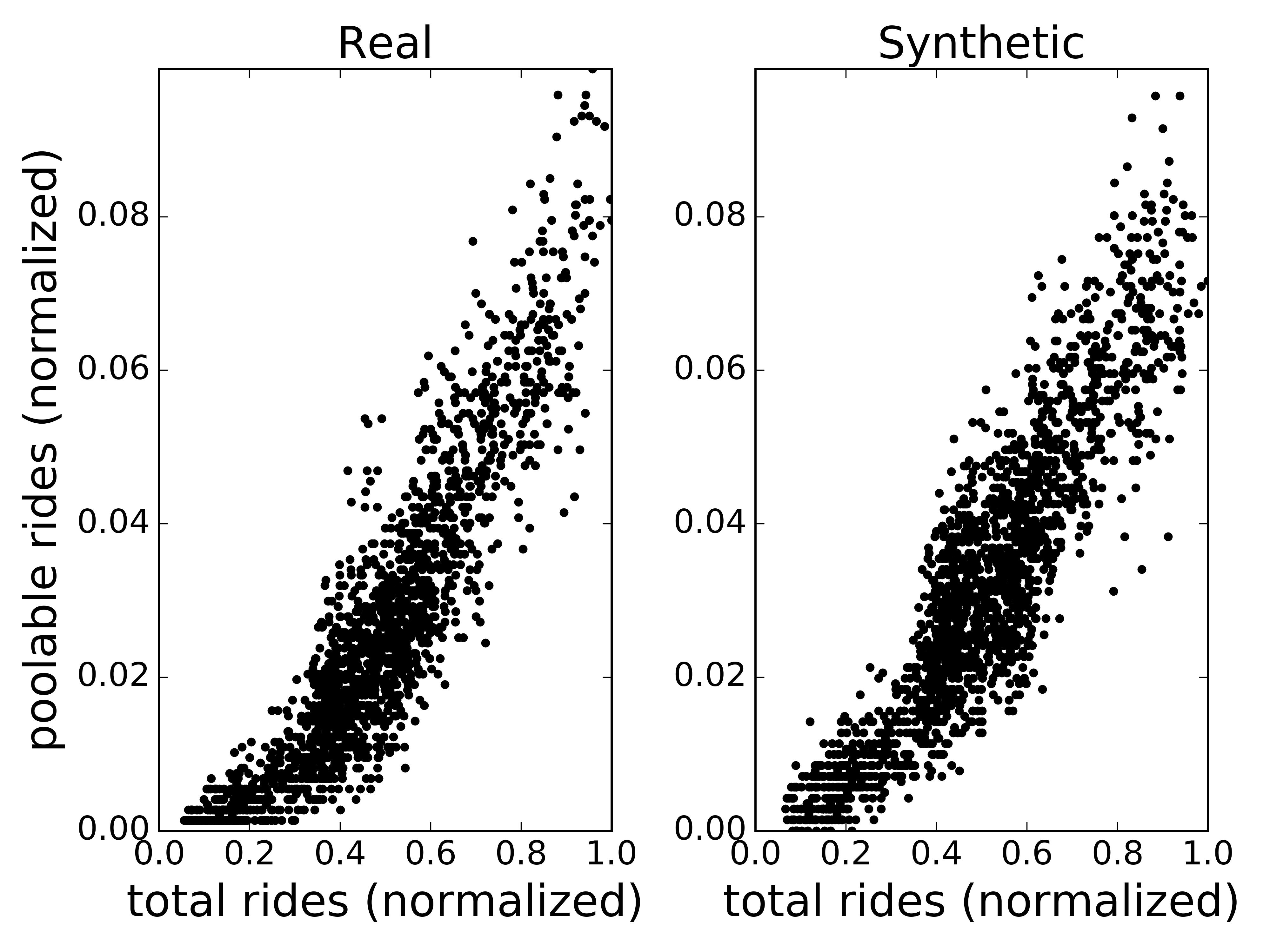}
    \caption*{New York \newline Real LS fit: slope=$0.088$ \newline Synthetic LS fit: slope$=0.088$}
    \label{fig:new_york_poolability}
\end{subfigure}%
\begin{subfigure}[t]{.5\linewidth}
  \centering
  \includegraphics[width=\linewidth, height=3.5cm]{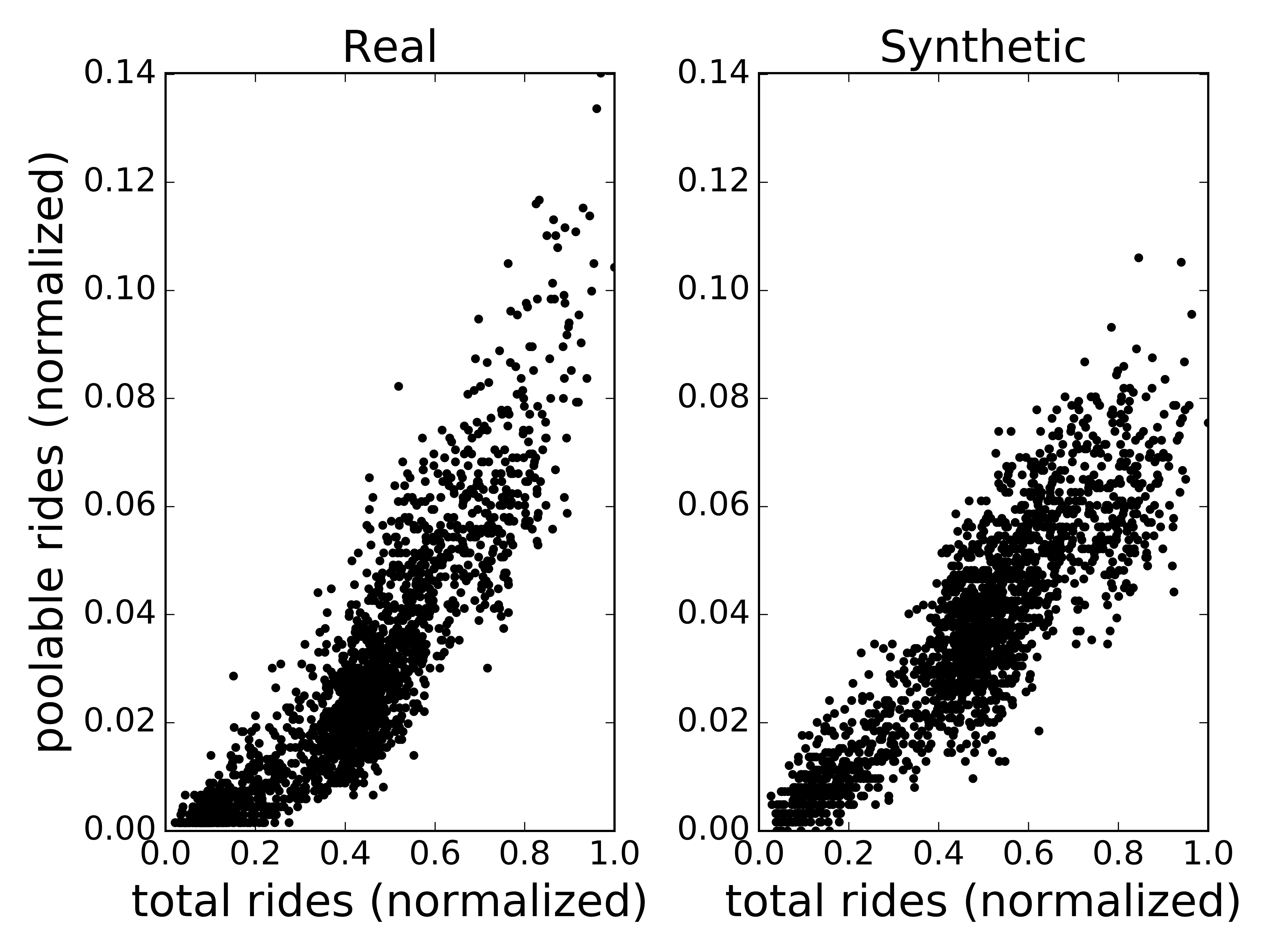}
    \caption*{San Francisco \newline Real LS fit: slope=$0.096$ \newline Synthetic LS fit: slope$=0.084$}
    \label{fig:san_francisco_poolability}
\end{subfigure}
\caption{Scatter plots of poolable rides vs. total rides for $2016$ 5-minute intervals in a week, based on the RRGs from real data (left) and the synthesized RRGs (right).}
\label{fig:poolability}
\end{figure}

\subsection{Comparison of Graph Models}
Figure~\ref{fig:edge_node} (bottom row) provides plots for the synthesized graph model with densification factors very similar to those from RRGs generated from real data in Figure~\ref{fig:edge_node} (top row). Figure~\ref{fig:real_and_random_nodes} shows the spatial distribution produced by Algorithms~\textsc{spatialPropGen} and \textsc{randomWalk} for the city of San Francisco. The plots show nodes (aggregation of points) in an RRG for a single time interval for both the real RRG (left) and the synthesized RRG (right). The nodes are more closely packed in the synthetic plot which is due to bias induced by the prior probability distribution using PoI density. Certain geographical areas in Figure~\ref{fig:san_francisco_2_synthetic} have no nodes in comparison to Figure~\ref{fig:san_francisco_2_real}, also due to the prior PoI density distribution being low in sparse areas.

\section{Ride Request Poolability}\label{sec:ride_request_poolability} 
\begin{figure*}[!t]
\captionsetup{justification=centering}
\centering
\begin{subfigure}{.25\textwidth}
  \centering
    \includegraphics[width=\linewidth]{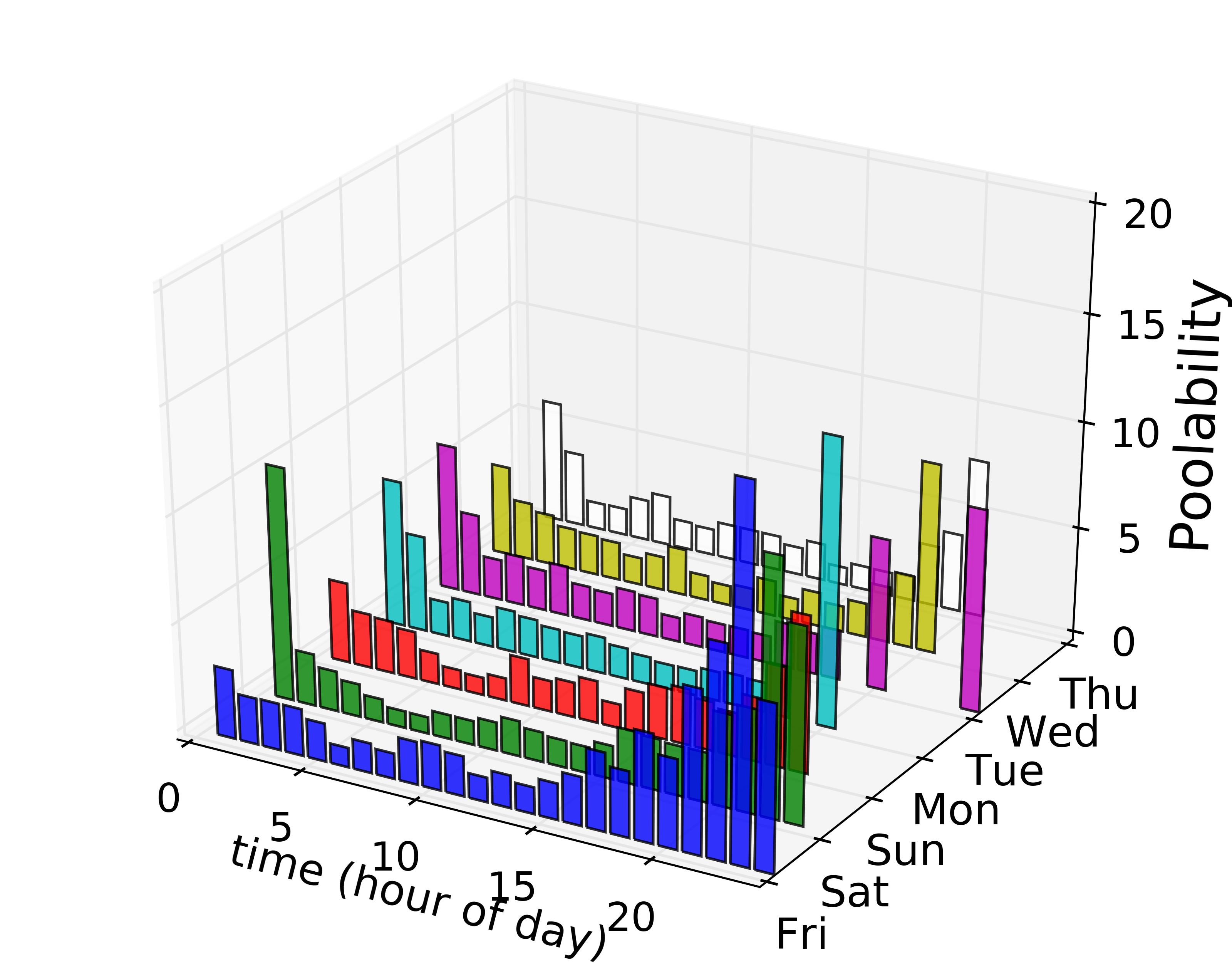}
    \caption*{Hyderabad}
    \label{fig:paris_pp_fixed}
\end{subfigure}%
\begin{subfigure}{.25\textwidth}
  \centering
    \includegraphics[width=\linewidth]{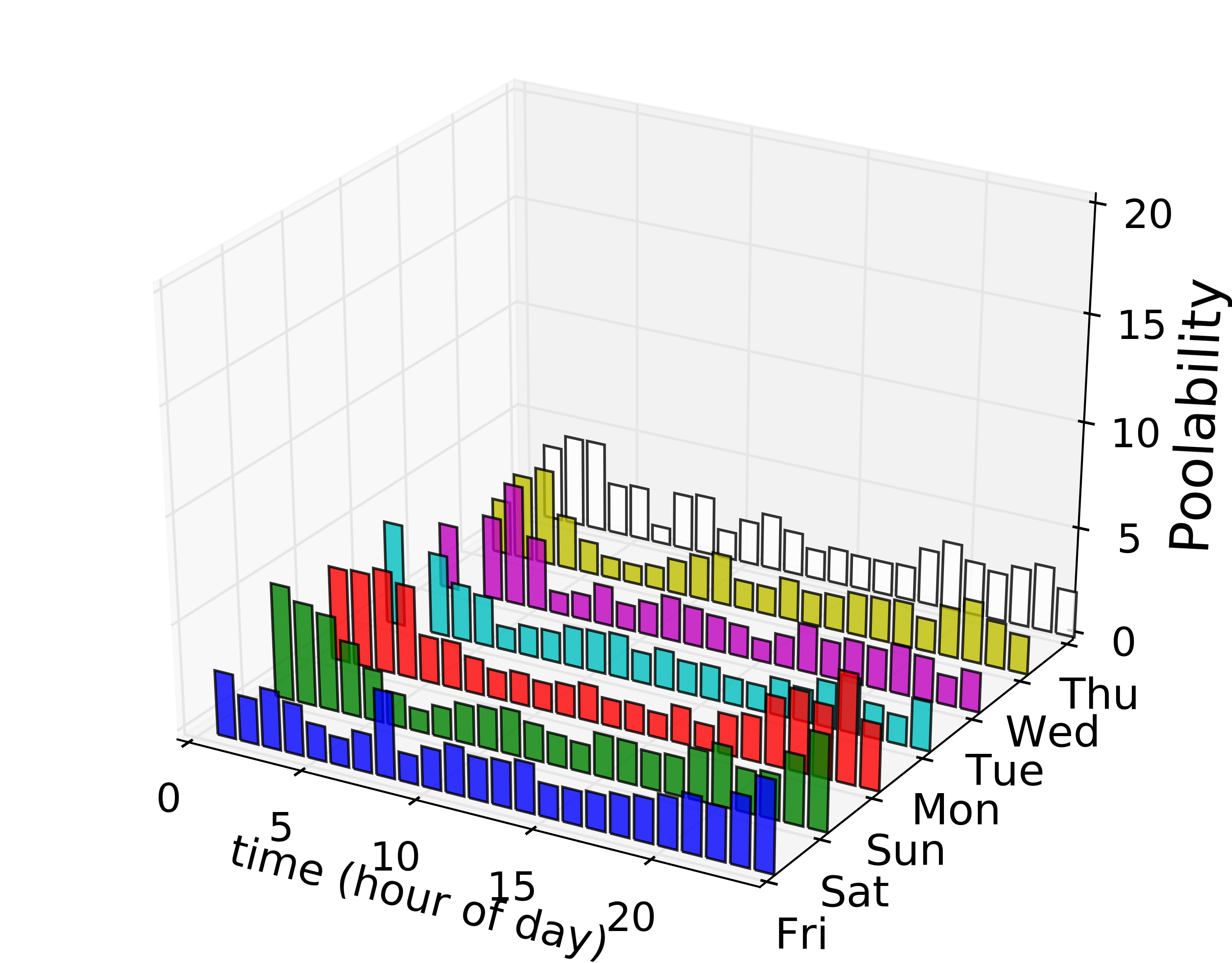}
    \caption*{Paris}
    \label{fig:hyderabad_pp_fixed}
\end{subfigure}%
\begin{subfigure}{.25\textwidth}
  \centering
    \includegraphics[width=\linewidth]{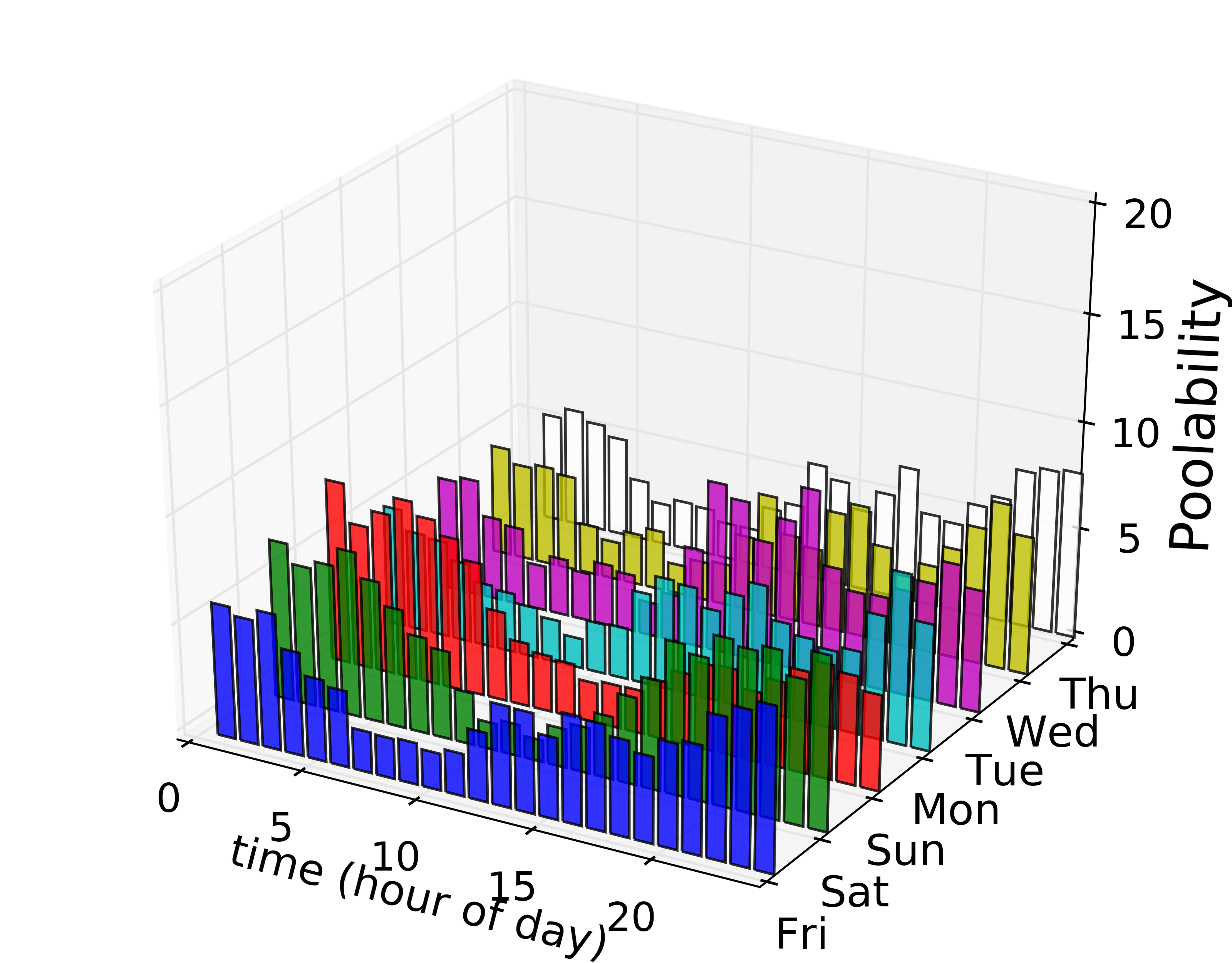}
    \caption*{New York}
    \label{fig:new_york_pp_fixed}
\end{subfigure}%
\begin{subfigure}{.25\textwidth}
  \centering
    \includegraphics[width=\linewidth]{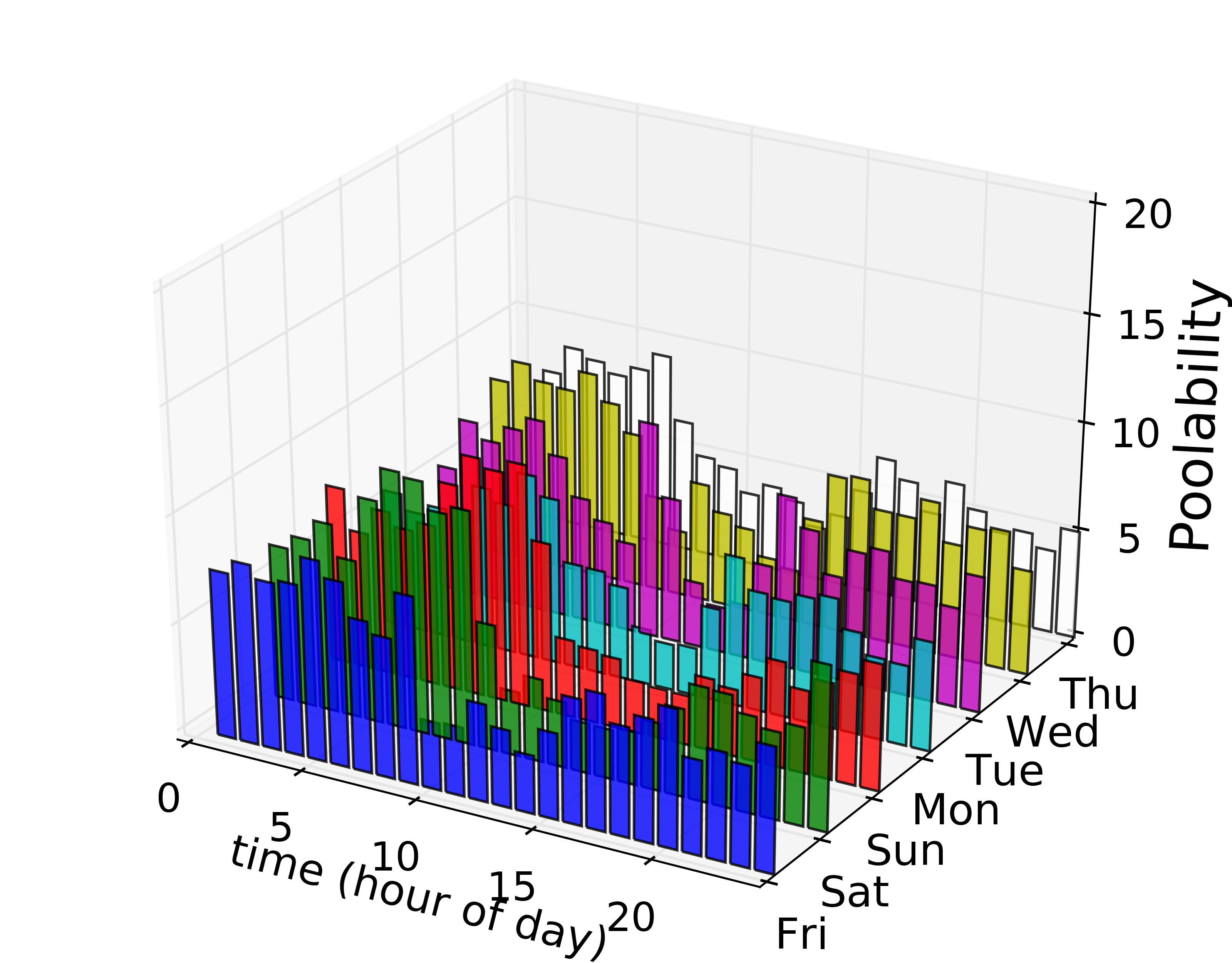}
    \caption*{San Francisco}
    \label{fig:san_francisco_pp_fixed}
\end{subfigure}
\caption{Poolability metric for four cities for a week of data with $\Delta t=5min, \Delta s=100m, \Delta d=1000m$. Time is in GMT.}
\label{fig:real_poolability}
\end{figure*}

Recently, ride-sharing services have started offering the option of ride pooling by matching similar ride requests in real time. Such ride or rider matching problem is similar to the well studied combinatorial problem commonly referred to as the \textit{Vehicle Routing Problem}~\cite{laporte1992vehicle}. An interesting and powerful approach to maximize pooling and minimize costs is by advanced scheduling, wherein a rider provides a time period in the future for pick-up, and the ride-sharing service performs matching within the time period. 

In this work, we instead consider the potential for on-demand ride pooling, i.e. pooling rides not scheduled in advance. We first focus on assessing the potential of ride pooling based on historical data. We examine all the ride requests in a city and attempt to bundle ride requests within certain proximity constraints in both space and time. For example, we can pool ride requests initiated within a 5 minute window, with requesting locations less than 100m apart.

Consider a set of rides $P$ ordered by time of request, such that $\left\vert{P}\right\vert = p, p > 1$; the first ride to occur in time in $P$ is referred to as the  \textit{master} ride.  Then \textit{master} is \textit{poolable} with any request $\in P \setminus \left\{ {master}\right\}$ if the following constraints are satisfied: 
\begin{enumerate}
\itemsep0em 
\item both ride requests are requested within $\Delta t$ minutes.
\item source locations of both requests are within $\Delta s$ meters radius.
\item destination locations of both requests are within $\Delta d$ meters radius.
\end{enumerate}
All rides in such proximity with \textit{master}, and \textit{master} itself are removed from $P$, and the above steps are repeated with a new \textit{master} being the next earliest ride request in $P$. Any requests that remain unmatched are considered not \textit{poolable}.

\begin{table}[H]
\begin{center}
 \begin{tabular}{c|c|c|c} 
 \textit{City} & \textit{Mean} & \textit{Minimum} & \textit{Maximum} \\ [0.5ex] 
 \hline 
 Hyderabad & 2.23 & 0.84 & 7.41 \\
 Paris & 2.39 & 0.79 & 4.22 \\
 New York & 4.48 & 1.70 & 7.84 \\
 San Francisco & 5.48 & 2.50 & 9.16 
\end{tabular}
\end{center}
\caption{Overall mean, minimum, and maximum poolability for four cities for a week of data with $\Delta t=5min, \Delta s=100m, \Delta d=1000m$}
\label{tab:poolability}
\end{table}

We define \textit{poolability} as the percentage of rides that can be pooled. 
In Figure~\ref{fig:real_poolability} we plot the \textit{poolability} (z-axis) for four cities. The poolability data is shown for each day of the week. Summary of poolability metric for four cities is provided in Table~\ref{tab:poolability}. Poolability in Hyderabad shows maximum variability with minimum of $0.84$, and maximum of $7.41$.

Paris and San Francisco exhibit quite different degrees of poolability. San Francisco consistently exhibits higher poolability, with a consistent daily pattern for weekdays. For each day there are two time periods, matching the morning and evening rush hours, that exhibit significantly higher poolability. We suspect the key difference between Paris and San Francisco is due to the topology and terrain of the two cities. This is a very interesting area for future research.

\subsection{Ride Poolability Attributes}
We also observe that the poolability of a city is directly correlated with its densification factor. Cities with higher $\alpha$ always exhibit higher poolability. Comparing Figure~\ref{fig:real_poolability} with the top row of Figure~\ref{fig:edge_node} we see that $\alpha = 1.031, 1.054, 1.098, 1.104$, for Hyderabad, Paris, New York, and San Francisco, respectively. This ordering matches exactly the ordering of poolability in Figure~\ref{fig:real_poolability} and Table~\ref{tab:poolability} (mean poolability). 

\subsection{Synthesized RRG Poolability}
Figure~\ref{fig:poolability} provides a comparison of \textit{poolability} obtained using real data and randomly generated data. The slope of the straight line fitted to real and synthetic plots suggests that the synthesized graph model is a relatively good fit to the real \textit{poolability}. There are instances where the synthesized version over or under predicts poolability. This is most likely due to the lack of spatial information of ride requests distributed over time. Node density information from OSM is not dynamic relative to the time of day, hence there is a bias towards generating points in high PoI density regions which may not hold for consecutive time intervals.  

\begin{figure}[!t]
  \centering
    \includegraphics[width=\linewidth]{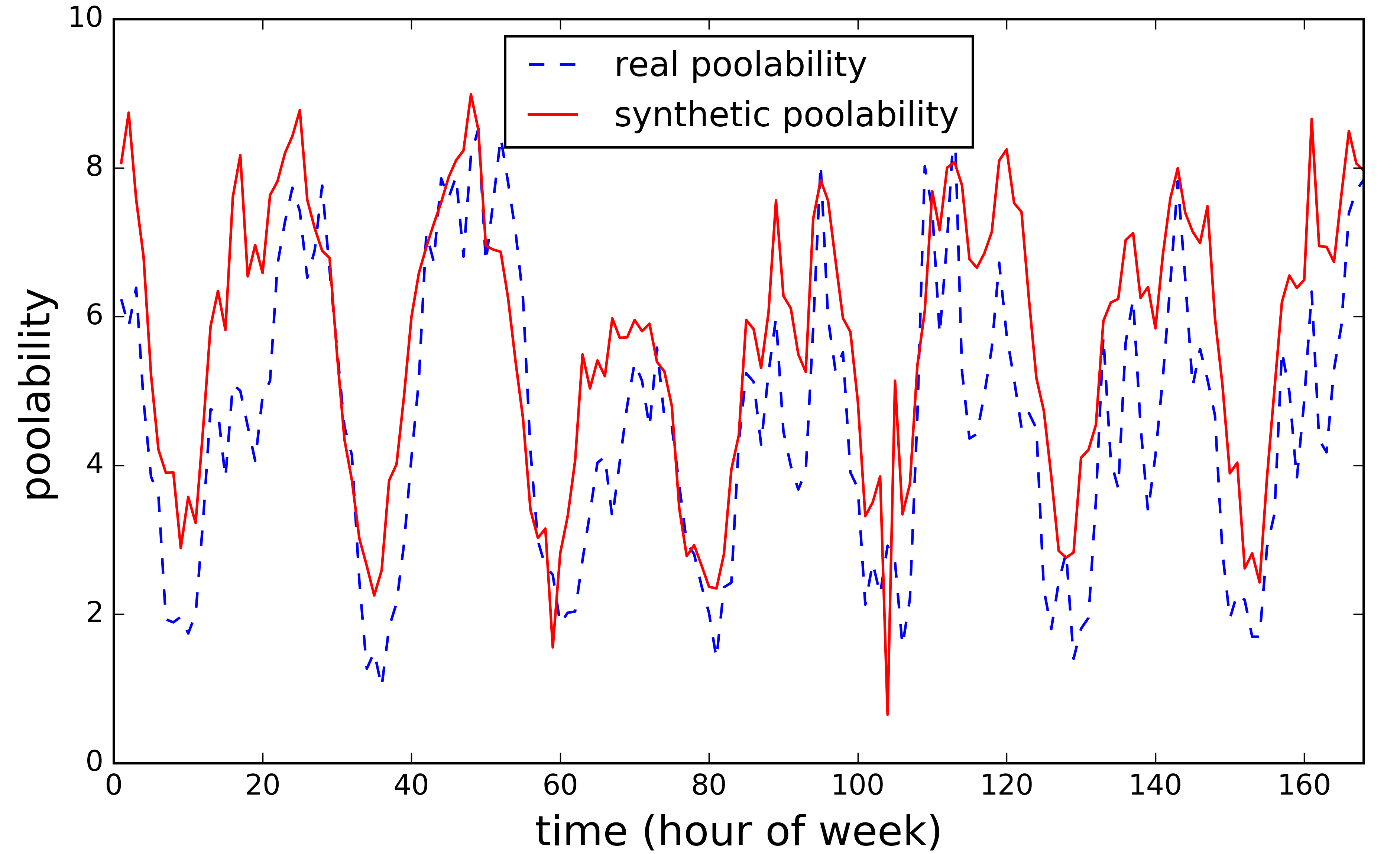}
    \caption{Comparison of \textit{Poolability} generated by synthetic (red line) and real (dotted blue line) models for New York. RMSE: 1.54, abs. delta min=0.02, abs. delta max=3.17}
    \label{fig:poolability_time_series}
\end{figure}

Figure~\ref{fig:poolability_time_series} shows how the \textit{poolability} metric varies over $168$ hours (starting with 8pm on Friday) of a typical week. The synthetic model captures the temporal variations and matches well with the \textit{poolability} from the real ride request data. For the synthetic model, the total number of ride requests were kept equal to the number of ride requests in real data. The first three peaks in the plot depict evening hours for Friday, Saturday, and Sunday which on average are higher than the remaining four peaks in the plot. 

\section{Conclusion}\label{sec:conclusion}
The emergence of ride sharing services and the availability of extensive data from such services is creating unprecedented opportunities for: doing large scale data analytics on urban transportation; gaining new insights on human mobility; and facilitating new public services for societal benefit. This work is an initial attempt at this. The key contributions of this paper include: 
\begin{itemize}
\itemsep0em 
\item Based on extensive real world data, we introduce a space-time framework for modeling ride requests in a city, and the notion and analysis of ride poolability. 
\item We introduce a space-time graph model for modeling ride requests in a city and show that these graphs exhibit power law densification as they evolve in time. 
\item Based on the densification power law, we show that the pattern of ride requests and ride poolability for a city can be concisely characterized by the densification factor of its ride request graphs.
\item We further show that the degree of ride poolability of a city is directly correlated to the densification factor of its ride request graphs.
\item Using previous work, we show the space-time ride request graph model for a city can be automatically generated. 
\item We further show the attributes of the generated synthetic graphs  match quite well the attributes of graphs extracted from real ride request data. 
\end{itemize}
We have only scratched the surface in this paper. There are many promising avenues for further research. Some open research questions include: 
\begin{enumerate}
\itemsep0em 
\item If the ride pooling proximity constraints, both temporal and spatial can be relaxed,  is it possible to significantly improve  ride poolability?
\item Can the temporal and spatial variation of ride poolability be leveraged to create intelligent ride pooling algorithms?
\item Is it possible to significantly reduce the number of vehicles needed on the road through aggressive ride pooling?
\item Can we rigorously characterize the relationship between the degree of ride poolability and the densification factor of ride request graphs?
\item Is it possible to use insights from historical ride request data for real-time traffic congestion prediction and potentially alleviation?
\item Comparison of RRG generation with existing graph generators over space and time.
\end{enumerate}
\emph{Acknowledgements:} We are grateful to Peter Frazier and Jon Petersen for many useful discussions. We also thank the anonymous referees for their helpful comments.
\bibliographystyle{aaai}
\bibliography{refs}
\end{document}